\documentclass[10pt,twocolumn,letterpaper]{article}

\usepackage{cvpr}              

\usepackage{multirow}

\usepackage[table]{xcolor}

\definecolor{baselinebg}{RGB}{243,243,243}

\usepackage{booktabs}
\usepackage{graphicx}
\definecolor{vinca}{HTML}{7B7FD8}
\colorlet{oursbg}{vinca!17}
\usepackage{placeins}
\usepackage{tabularx}

\definecolor{cvprblue}{rgb}{0.21,0.49,0.74}
\usepackage[pagebackref,breaklinks,colorlinks,allcolors=cvprblue]{hyperref}

\def\confName{CVPR}
\def\confYear{2027}

\title{\raisebox{-0.1em}{\includegraphics[height=1.0em]{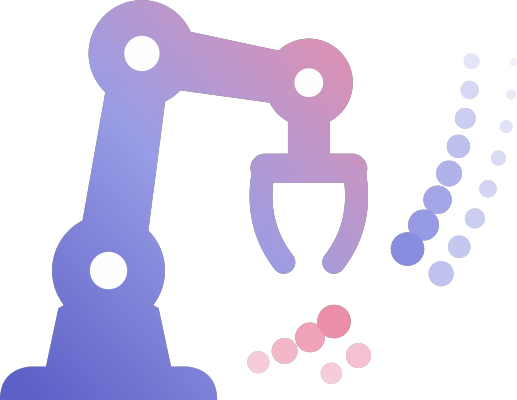}Event-WAM}: Learning to Act with Event-Derived Futures}

\author{
Jongseong Bae\thanks{Equal contribution.} \quad
Hyunwoo Kim\footnotemark[1] \quad
Yeongin Lee\footnotemark[1] \quad
Eubin So \quad
Ha Young Kim\thanks{Corresponding author.}\\
Yonsei University\\
{\tt\small
\{js.bae,hw.kim,zeroin,ebso5062,hayoung.kim\}@yonsei.ac.kr}
}

\begin{document}
\maketitle

\begin{abstract}
World Action Models (WAMs) leverage future visual modeling for robotic action learning, with recent studies exploring alternative or complementary objectives beyond conventional RGB prediction. We introduce \textbf{Event-WAM}, which incorporates event-derived visual changes as auxiliary future supervision alongside RGB. Event-WAM constructs temporally structured event representations from RGB demonstrations and encodes them using a frozen pretrained RGB video VAE, without training an event-specific VAE. Joint RGB--Event latent flow matching provides appearance- and change-oriented supervision without requiring event observations or explicit future generation at inference. Across simulated benchmarks, Event-WAM maintains comparable performance under standard conditions, while robustness gains under perturbations are benchmark-dependent, including improvements on LIBERO-Plus. Real-world evaluations further demonstrate improved manipulation success under both standard and unseen perturbation conditions. Further analyses show that policy improvements are not consistently accompanied by higher future prediction fidelity, suggesting the potential of change-oriented supervision for action learning.
\end{abstract}

\section{Introduction}
\label{sec:intro}

Recent advances in visual world modeling have established World Action Models (WAMs) as a promising approach to robotic manipulation. By jointly learning future visual dynamics and robot actions, WAMs leverage predictive supervision to support action generation. Early approaches~\cite{unipi,dreamzero} rely on iterative future video generation at inference, whereas Fast-WAM~\cite{fastwam} demonstrates that future modeling can benefit action learning without generating future video. More recently, Simple-WAM~\cite{zhou2026makes} shows that a single forward pass over fully noised future tokens can effectively condition action generation.
Together, these advances shift attention from whether future videos must be generated at inference to how training-time future modeling shapes action-relevant representations.

Alongside this progress, an important question concerns which aspects of the future should be modeled to support action learning. Conventional RGB-based future modeling provides rich supervision for both scene appearance and temporal evolution, but dense visual prediction does not explicitly emphasize dynamic changes over persistent scene content~\cite{dcwam,yan2026deltawam,dreamwam}. Recent studies have explored alternative or complementary future modeling objectives, including change-centric targets based on visual deltas, residuals, and temporal differences~\cite{yan2026deltawam,wu2026delta,dcwam}, as well as latent and multimodal future representations involving dynamics, geometry, and semantics~\cite{lawam,dreamwam,hung2026modality}, reflecting ongoing efforts to identify effective supervision targets for action learning.

In this context, we consider event-based representations as an alternative form of visual-change supervision for WAMs. 
Events~\cite{event_survey} encode pixel-wise log-intensity changes through contrast thresholding, preserving polarity and temporal structure while suppressing unchanged content. While RGB frames provide dense appearance and static scene context, events emphasize sparse, temporally structured visual changes, complementing RGB appearance with change-oriented cues~\cite{jiang2024evrgbhand,zhou2024eventfusion}. 
These properties have encouraged the joint use of RGB and events across a wide range of vision tasks~\cite{jiang2024evrgbhand,zhou2024eventfusion,xu2024dmr,kong2025dessert}, yet whether RGB-derived event representations can provide effective future supervision for action learning, rather than serving as additional sensory inputs, remains largely unexplored in WAMs.

To this end, we introduce \textbf{Event-WAM}, which integrates event-derived visual-change supervision with RGB future modeling for world-action learning. 
We construct temporally binned, polarity-separated event representations from existing RGB demonstrations and encode them using the same frozen pretrained RGB video VAE, mapping both RGB and event representations into a common latent space without training an event-specific encoder.
Within a shared video backbone, Event-WAM jointly models RGB and event latents through parallel flow-matching objectives with independently sampled noise, combining appearance-oriented supervision with temporally organized visual-change signals. 
Following Simple-WAM's inference strategy, our approach incorporates both future modeling objectives into action learning without requiring event observations or explicit future generation at inference.

We evaluate Event-WAM on simulated manipulation benchmarks, including LIBERO~\cite{libero}, LIBERO-Plus~\cite{fei2025libero}, and RoboTwin 2.0~\cite{robotwin2}, as well as four real-world manipulation tasks using an SO-101 robotic arm~\cite{so101}.
Event-WAM maintains comparable performance under standard simulated conditions and improves robustness on LIBERO-Plus and RoboTwin 2.0-Plus, but not under RoboTwin C2R randomization.
In real-world manipulation, it further improves success under both standard and unseen perturbation conditions.
We investigate RGB and event-based future modeling through loss-weighting ablations, single-target objectives, and analyses of future prediction fidelity. The results show that improvements in action performance are not consistently accompanied by higher future prediction quality.
Since event representations are derived from RGB observations, these findings suggest that their benefit may arise from emphasizing action-relevant visual changes during future modeling rather than introducing additional sensory information. This interpretation is consistent with the limited gains observed in standard simulated settings and the more pronounced improvements on LIBERO-Plus and in real-world manipulation.

Our main contributions are summarized as follows:

\begin{itemize}

\item We investigate event-derived visual changes as complementary generative future targets alongside RGB in WAMs, examining their effects on robotic action learning and policy generalization.

\item We introduce Event-WAM, which jointly models RGB and event-derived futures through flow matching within a shared pretrained backbone, combining appearance- and change-oriented supervision for action learning.

\item We evaluate Event-WAM across simulated and real-world settings, demonstrating improved robustness under perturbations and higher success on physical robot tasks, with further analyses of supervision targets and of how prediction fidelity relates to action performance.
\end{itemize}

\section{Related Work}
\label{sec:related}

\subsection{World Action Models}
WAMs couple future-state modeling with robot action generation, often leveraging pretrained video diffusion or flow-matching backbones~\cite{unipi,uva,dreamzero,fastwam}.
Fast-WAM~\cite{fastwam} demonstrates that training-time future modeling can benefit action learning without requiring future video generation at inference.
More recently, ForeWAM~\cite{forewam} utilizes key--value representations of noise-initialized future slots for efficient action conditioning, while Simple-WAM~\cite{zhou2026makes} shows that a single forward pass over fully noised future tokens retains much of the generalization benefit of explicit future modeling.
Faster-WAM~\cite{fasterwam_future} and BridgeWAM~\cite{bridgewam} further improve inference efficiency through sparse video--action interactions and compact latent bridge queries, respectively.
OpenWAM-$\alpha$~\cite{openwam} explores scalable world--action pretraining using egocentric and robot data.

Recent studies explore alternative and complementary future modeling objectives to better capture visual dynamics.
Some approaches predict structured or compact future representations~\cite{hung2026modality,lawam}, while others emphasize visual changes through frame differences and feature deltas~\cite{yan2026deltawam,jiang2026deltawam,wu2026delta}.
EA-WM~\cite{eawm} uses VAE-encoded absolute RGB frame differences to supervise intermediate action--video fusion representations, whereas DC-WAM~\cite{dcwam} emphasizes interaction-induced dynamics through temporal-difference supervision and dynamic attention routing.
Geometric and multimodal approaches further expand future supervision: 4D-WAM~\cite{4dwam} aligns representations with 3D trajectory-field features, PrismWAM~\cite{prismwam} separates RGB-based semantic planning from depth-based physical rollout, and DreamWAM~\cite{dreamwam} jointly models RGB and optical flow with auxiliary geometric and semantic objectives.
MT-WAM~\cite{mtwam} additionally uses future motion trajectories and visual features to shape one-pass predictive representations for action generation.
In contrast, Event-WAM incorporates event-derived representations as complementary generative future targets jointly modeled with RGB through flow matching, using shared representations.

\begin{figure*}[hbt!]
    \centering
    \IfFileExists{figures/fig1-3.pdf}{%
        \includegraphics[width=\textwidth]{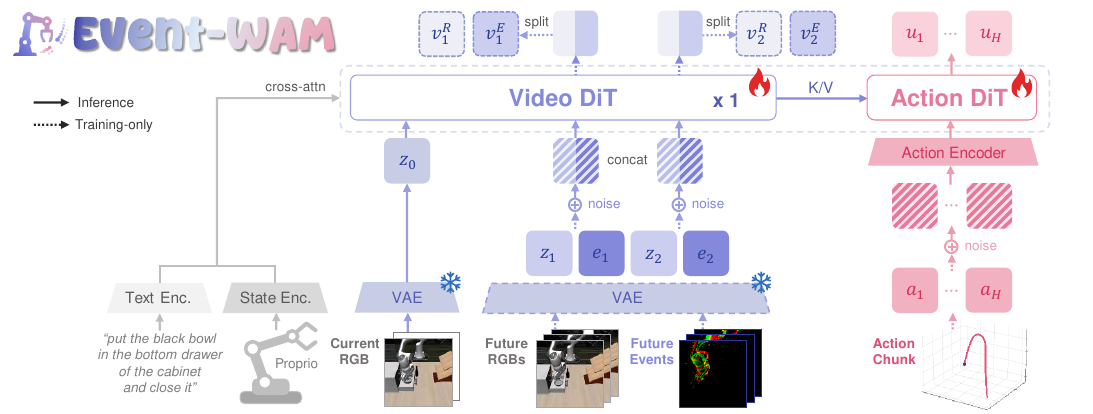}%
    }{%
    \IfFileExists{figures/fig1-3.pdf}{%
        \includegraphics[width=\textwidth]{figures/fig1-3.pdf}%
    }{%
        \fbox{\parbox[c][1.65in][c]{0.94\textwidth}{\centering
        ARCHITECTURE FIGURE PLACEHOLDER\\[4pt]
        (a) Simple-WAM \quad (b) Event-WAM-Joint \quad (c) Event-WAM-Residual
        }}%
    }}
    \caption{\textbf{Overview of Event-WAM.}
    A shared pretrained RGB video VAE encodes RGB and event sequences into latents, whose future components are jointly modeled via flow matching.
    At inference, fully noised future tokens undergo a single video-model pass, and the resulting KV features condition iterative action denoising without future generation.
    Dotted arrows indicate training-only pathways.}
    \label{fig:overview}
\vspace{-3.0mm}
\end{figure*}

\subsection{Event-Based Vision and Robot Learning}
Event cameras~\cite{event_survey} capture asynchronous pixel-wise log-intensity changes, retaining polarity and fine-grained temporal information.
Unlike RGB frames, which provide dense scene appearance and spatial context, events emphasize brightness transitions while suppressing unchanged content.
These complementary characteristics have motivated RGB--Event fusion across various visual perception tasks~\cite{jiang2024evrgbhand,zhou2024eventfusion,xu2024dmr,kong2025dessert}.
Given the limited availability of large-scale event-camera datasets, event simulation methods~\cite{esim,v2e,eventgan} enable the synthesis of event streams from conventional RGB videos without event cameras, facilitating broader research in event-based vision.

Beyond perception, event representations have also been explored for future motion modeling and robotic policy learning.
E-Motion~\cite{emotion} investigates future motion modeling from event sequences.
In robotic learning, SEBVS~\cite{sebvs} explores simulated event observations for navigation and manipulation, while E-VLA~\cite{evla} and Event-VLA~\cite{eventvla} incorporate event observations into vision-language-action models.
ECHO~\cite{wang2026echo} further combines wrist-mounted event observations with event-foresight queries in a latent world-action model.
Whereas these robotic approaches require event observations at inference, Event-WAM uses RGB-derived events solely as training-time future targets alongside RGB supervision, without additional event sensing.

\section{Method}
\label{sec:method}
Event-WAM integrates event-derived visual changes into RGB-based world-action learning as complementary future supervision. Building on the RGB-only formulation of Simple-WAM~\cite{zhou2026makes}, our approach jointly models future RGB and event latents through shared-backbone flow matching, while retaining single-pass future-token conditioning followed by iterative action denoising at inference. Figure~\ref{fig:overview} illustrates the overall architecture. We first review the preliminaries of WAMs and event cameras (Sec.~\ref{sec:common_wam}), followed by event-derived future representation and joint RGB--Event modeling within Event-WAM (Sec.~\ref{sec:event_wam}).

\subsection{Preliminaries}
\label{sec:common_wam}

\paragraph{World action models.}
World Action Models (WAMs) couple future visual modeling with robot action generation through a video model and an action model~\cite{fastwam,zhou2026makes}.
Given an RGB observation $I_0$, a proprioceptive state $\mathbf q$, and a language instruction $\ell$, a WAM learns to predict an action chunk $\mathbf a$ by leveraging future visual representations.

In a typical RGB-based WAM formulation, an RGB sequence $\mathbf I$ containing the observation and future frames is encoded using a frozen causal video VAE $\mathcal E$~\cite{wan2025}, yielding the observation latent $\mathbf z_0$ and future latents $\mathbf z_F$.
In the flow-matching formulation~\cite{lipman2022flow,liu2022flow}, the future latents are interpolated with Gaussian noise as $\mathbf x_\sigma=(1-\sigma)\mathbf z_F+\sigma\boldsymbol\epsilon$, where $\boldsymbol\epsilon\sim\mathcal N(\mathbf 0,\mathbf I)$. The video model processes the resulting latent sequence and produces conditioning features:
\begin{equation}
    \mathbf Y_\sigma=[\mathbf z_0;\mathbf x_\sigma],
    \qquad
    \mathbf h_\theta=
    \operatorname{KV}_\theta(\mathbf Y_\sigma,\sigma;\ell,\mathbf q),
    \label{eq:wam_formulation}
\end{equation}
where $\mathbf h_\theta$ denotes the attention keys and values extracted from the video model. 
The video model $\mathbf v_\theta$ predicts the future-latent velocity, while the action model $\mathbf u_\psi$ predicts the action velocity conditioned on $\mathbf h_\theta$, $\ell$, and $\mathbf q$.
Their training objectives are
\begin{equation}
\begin{aligned}
\mathcal L_{\mathrm{RGB}}
&=\mathbb E_{\sigma,\boldsymbol\epsilon}
\left[\omega(\sigma)
\left\|\mathbf v_\theta(\mathbf Y_\sigma,\sigma;\ell,\mathbf q)
-(\boldsymbol\epsilon-\mathbf z_F)\right\|^2\right],\\
\mathcal L_{\mathrm{act}}
&=\mathbb E_{\sigma,\boldsymbol\epsilon,\tau,\boldsymbol\epsilon_a}
\left[\omega_a(\tau)
\left\|\mathbf u_\psi(
\mathbf a_\tau,\tau;\mathbf h_\theta,\ell,\mathbf q)
-(\boldsymbol\epsilon_a-\mathbf a)\right\|^2\right],
\end{aligned}
\label{eq:wam_losses}
\end{equation}
where $\mathbf a_\tau=(1-\tau)\mathbf a+\tau\boldsymbol\epsilon_a$, $\boldsymbol\epsilon_a\sim\mathcal N(\mathbf 0,\mathbf I)$, and $\omega$ and $\omega_a$ denote timestep weights. 
The RGB-based WAM is jointly optimized with $\mathcal L_{\mathrm{WAM}}=\lambda_{\mathrm{RGB}}\mathcal L_{\mathrm{RGB}}+\mathcal L_{\mathrm{act}}$. We adopt the mixed noise training strategy of Simple-WAM~\cite{zhou2026makes}, setting $\sigma=1$ with probability $p_1$ and otherwise sampling from the noise schedule to align training with single-pass future-token conditioning at test time.

At inference, early WAMs iteratively denoise future latents $\mathbf x_\sigma$, whereas Fast-WAM~\cite{fastwam} omits future tokens entirely. We follow Simple-WAM~\cite{zhou2026makes}, which retains future tokens at $\sigma=1$ and computes their representations in a single video-model pass:
\begin{equation}
    \mathbf h_\theta^{\mathrm{inf}}
    =\operatorname{KV}_\theta(
    [\mathbf z_0;\boldsymbol\epsilon],1;\ell,\mathbf q).
    \label{eq:single_pass}
\end{equation}
The resulting KV cache remains fixed while
$\mathbf u_\psi(\mathbf a_\tau,\tau;
\mathbf h_\theta^{\mathrm{inf}},\ell,\mathbf q)$
iteratively denoises the action chunk, without explicitly generating future visual latents.

\paragraph{Event cameras.}
Event cameras asynchronously capture changes in log intensity as a stream of events $\mathcal S=\{(x_k,y_k,t_k,p_k)\}_{k=1}^{N}$, where $(x_k,y_k)$ denotes the pixel location, $t_k$ the timestamp, and $p_k\in\{+1,-1\}$ the polarity~\cite{event_survey}.
An event is triggered when the change in log intensity since the last event at that pixel reaches a contrast threshold, with positive (ON, $+1$) and negative (OFF, $-1$) polarities indicating brightness increases and decreases, respectively.
Unlike conventional RGB frames, event streams emphasize temporal brightness changes, while regions with little intensity change produce few or no events.
In this work, we synthesize event streams from RGB demonstration videos using the video-to-event simulator v2e~\cite{v2e}, without requiring additional event-camera observations.

\subsection{Event-WAM}
\label{sec:event_wam}

\paragraph{Event-derived future representation.}
Given the simulated event stream $\mathcal S$, we partition it into $B$ consecutive temporal bins $\{\mathcal T_b\}_{b=1}^{B}$ aligned with the RGB prediction horizon. For each bin $b$, we accumulate the number of events at each pixel separately for the two polarities:
\begin{equation}
    n_b^p(x,y)
    =\big|\{k:(x_k,y_k)=(x,y),
    \ t_k\in\mathcal T_b,\ p_k=p\}\big|,
    \label{eq:event_count}
\end{equation}
where $p\in\{+1,-1\}$ denotes the event polarity. The polarity-separated event counts capture the spatial distribution and polarity of brightness changes within each bin, while retaining coarse temporal structure across bins.

To obtain event-derived future latents without training a separate event-specific VAE, we convert the polarity-separated counts into three-channel pseudo-RGB frames compatible with the frozen RGB video encoder $\mathcal E$. We first apply a logarithmic transform to compress the dynamic range of event counts:
\begin{equation}
\begin{aligned}
    P_b &= \log(1+n_b^+),\\
    N_b &= \log(1+n_b^-).
\end{aligned}
\label{eq:event_log_counts}
\end{equation}
Each event frame is then constructed as
\begin{equation}
    E_b
    =\operatorname{clip}_{[-1,1]}
    \left(
    \frac{1}{\gamma_E}
    [2P_b,\,2N_b,\,P_b+N_b]-1
    \right),
    \label{eq:pseudo_rgb}
\end{equation}
where $\gamma_E$ is a normalization scale estimated from the training data. The first two channels encode ON and OFF log counts, respectively, while the third represents their sum.

To align the event representations with $\mathbf z_F$, we prepend an event-free reference frame, represented by $-1$ in each channel under Eq.~\eqref{eq:pseudo_rgb}, and arrange the event frames to match the temporal layout of the RGB VAE inputs:
\begin{equation}
    \mathbf E=[E_0;E_1;\ldots;E_B],
    \qquad E_0=-\mathbf 1.
    \label{eq:event_sequence}
\end{equation}
The event sequence is encoded using the same frozen VAE:
\begin{equation}
    [\mathbf e_0;\mathbf e_F]
    =\mathcal E(\mathbf E),
    \label{eq:event_latents}
\end{equation}
where $\mathbf e_0$ denotes the reference latent, which is discarded. The retained future event latents $\mathbf e_F$ match the temporal layout and dimensionality of the RGB future latents $\mathbf z_F$, enabling their joint use as future modeling targets.

\paragraph{Joint RGB--Event modeling.}
Event-WAM jointly models future RGB and event latents through a shared video model. We concatenate the RGB future latents $\mathbf z_F$ and event future latents $\mathbf e_F$ along the channel dimension, while retaining RGB-only observation conditioning:

\begin{equation}
\begin{aligned}
    \mathbf f_F &= \mathbf z_F \oplus \mathbf e_F,\\
    \mathbf f_0 &= \mathbf z_0 \oplus \mathbf 0,
\end{aligned}
\label{eq:joint_latents}
\end{equation}
where $\oplus$ denotes channel-wise concatenation. Accordingly, we expand the input and output projections of the pretrained video model from $C$ to $2C$ channels. We initialize the RGB components with their pretrained weights, zero-initialize the additional event input weights, and randomly initialize the event output weights.

For joint flow matching, we independently sample Gaussian noise for the RGB and event latents while sharing the same noise level $\sigma$:
\begin{equation}
\begin{aligned}
    \mathbf x_\sigma^R
    &=(1-\sigma)\mathbf z_F
    +\sigma\boldsymbol\epsilon_R,\\
    \mathbf x_\sigma^E
    &=(1-\sigma)\mathbf e_F
    +\sigma\boldsymbol\epsilon_E,
\end{aligned}
\label{eq:joint_noising}
\end{equation}
where $\boldsymbol\epsilon_R$ and $\boldsymbol\epsilon_E$ are independently sampled from standard Gaussian distributions. The joint noisy latents are concatenated and processed together with the observation latent to obtain the video conditioning features:
\begin{equation}
\begin{aligned}
    \mathbf x_\sigma^J
    &=\mathbf x_\sigma^R \oplus \mathbf x_\sigma^E,\\
    \mathbf Y_\sigma^J
    &=[\mathbf f_0;\mathbf x_\sigma^J],\\
    \mathbf h_\theta^J
    &=\operatorname{KV}_\theta(
    \mathbf Y_\sigma^J,\sigma;\ell,\mathbf q).
\end{aligned}
\label{eq:joint_formulation}
\end{equation}
The video model predicts velocities for both RGB and event latents from the joint representation.

The RGB flow-matching loss $\mathcal L_{\mathrm{RGB}}$ follows Eq.~\eqref{eq:wam_losses}, using the RGB output $\mathbf v_\theta^R(\mathbf Y_\sigma^J,\sigma;\ell,\mathbf q)$ and taking the expectation over both RGB and event noise.
The event flow-matching loss is defined as
\begin{equation}
\mathcal L_{\mathrm{Event}}
=\mathbb E_{\sigma,\boldsymbol\epsilon_R,\boldsymbol\epsilon_E}
\left[
\omega(\sigma)
\left\|
\mathbf v_\theta^E(\mathbf Y_\sigma^J,\sigma;\ell,\mathbf q)
-(\boldsymbol\epsilon_E-\mathbf e_F)
\right\|^2
\right],
\label{eq:event_fm_loss}
\end{equation}
where $\mathbf v_\theta^E$ denotes the event channels of the predicted velocity. The squared errors are averaged over channels, spatial dimensions, and valid future latent positions. Both RGB and event losses are applied under the mixed noise training strategy described in Sec.~\ref{sec:common_wam}, including at $\sigma=1$.

The action model is conditioned on the video features $\mathbf h_\theta^J$, together with the language instruction $\ell$ and proprioceptive state $\mathbf q$, and is trained using the action flow-matching objective in Eq.~\eqref{eq:wam_losses}. The overall training objective is
\begin{equation}
    \mathcal L_{\mathrm{Event\text{-}WAM}}
    =\lambda_{\mathrm{RGB}}\mathcal L_{\mathrm{RGB}}
    +\lambda_{\mathrm{Event}}\mathcal L_{\mathrm{Event}}
    +\mathcal L_{\mathrm{act}},
    \label{eq:joint_loss}
\end{equation}
where $\lambda_{\mathrm{RGB}}$ and $\lambda_{\mathrm{Event}}$ control the contributions of the two future modeling objectives. 
The joint video representation is optimized by both future modeling and action learning.

At inference, we follow the single-pass conditioning strategy in Eq.~\eqref{eq:single_pass}, initializing both future RGB and event latents with Gaussian noise:
\begin{equation}
    \mathbf h_\theta^{J,\mathrm{inf}}
    =\operatorname{KV}_\theta\left(
    [\mathbf f_0;
    \boldsymbol\epsilon_R \oplus \boldsymbol\epsilon_E],
    1;\ell,\mathbf q
    \right).
    \label{eq:joint_inference}
\end{equation}
The KV cache remains fixed throughout iterative action denoising. 
Thus, Event-WAM requires neither event observations nor future RGB or event generation at inference.

\begin{table}[t]
\centering
\small
\caption{\textbf{LIBERO and LIBERO-Plus comparison.}
Success rates (\%). Emb. PT.: action-labeled embodied pretraining;
Extra Vid. PT.: action-free robot/egocentric video pretraining beyond the pretrained backbone.
Original and Corrected denote LIBERO-Plus instruction protocols.
Corrected reference scores follow La-JEPA~\cite{lajepa}.
$\dagger$: 10,029 variants;
$\star$: unverified instruction handling.}
\label{tab:libero_corrected}
\setlength{\tabcolsep}{2.pt}
\resizebox{\columnwidth}{!}{%
\begin{tabular}{@{}lccccc@{}}
\toprule
\multirow{2}{*}{\textbf{Method}} &
\multirow{2}{*}{\shortstack[c]{\textbf{Emb.}\\\textbf{PT.}}} &
\multirow{2}{*}{\shortstack[c]{\textbf{Extra}\\\textbf{Vid. PT.}}} &
\multirow{2}{*}{\shortstack[c]{\textbf{LIBERO}\\\textbf{Avg.}}} &
\multicolumn{2}{c}{\textbf{LIBERO-Plus Avg.}} \\
\cmidrule(lr){5-6}
& & & & \textbf{Original} & \textbf{Corrected} \\
\midrule
\multicolumn{6}{@{}l}{\textcolor{gray}{\textit{VLA-based}}} \\ \midrule
$\pi_{0.5}$~\cite{pi05} & $\checkmark$ & $\times$ & 96.9 & 84.4 & 85.7$^{\dagger}$ \\
LingBot-VA~\cite{li2026causal} & $\checkmark$ & $\times$ & 98.5 & 69.5 & -- \\
\midrule
\multicolumn{6}{@{}l}{\textcolor{gray}{\textit{WAM-based}}} \\ \midrule
Fast-WAM~\cite{fastwam} & $\times$ & $\times$ & 97.6 & 51.5 & 70.8 \\
DC-WAM~\cite{dcwam} & $\times$ & $\times$ & 98.1 & 60.9 & -- \\
DreamWAM~\cite{dreamwam} & $\times$ & $\times$ & 98.9 & 74.6 & -- \\
PrismWAM~\cite{prismwam} & $\times$ & $\times$ & \textbf{99.4} & 79.4 & -- \\
SA-WAM$^{\star}$~\cite{sawam} & $\times$ & $\times$ & 98.4 & \textbf{86.6} & -- \\
ProWAM$^{\star}$~\cite{prowam} & $\times$ & $\checkmark$ & 98.8 & 85.8 & -- \\
Simple-WAM~\cite{zhou2026makes} & $\times$ & $\times$ & -- & 79.5 & -- \\
ImageWAM~\cite{imagewam} & $\times$ & $\times$ & 98.4 & 79.7 & 80.4 \\
Cosmos Policy~\cite{cosmospolicy} & $\times$ & $\times$ & 98.5 & 80.8$^{\dagger}$ & 84.9$^{\dagger}$ \\
NowWAM~\cite{nowwam} & $\times$ & $\times$ & 98.4 & 86.2 & 86.2 \\
La-JEPA S~\cite{lajepa} & $\times$ & $\times$ & 97.9 & 80.0 & 86.3 \\
La-JEPA L~\cite{lajepa} & $\times$ & $\times$ & 98.1 & 84.1 & \textbf{88.0} \\
\midrule
\multicolumn{6}{@{}l}{\textcolor{gray}{\textit{Our implementation}}} \\ \midrule
Simple-WAM & $\times$ & $\times$ & 98.9 & 78.9 & 84.0 \\
\rowcolor{oursbg}\textbf{Event-WAM} & $\times$ & $\times$ & 98.8 & 81.0 & 85.9 \\
\bottomrule
\end{tabular}%
}
\vspace{-4.0mm}
\end{table}
\section{Experiments}
\label{sec:experiments}

\subsection{Benchmarks and Evaluation Protocols.}
\label{sec:exp_setup}

We evaluate Event-WAM on LIBERO, LIBERO-Plus~\cite{libero,fei2025libero}, RoboTwin 2.0~\cite{robotwin2}, RoboTwin 2.0-Plus~\cite{zhang2026worldaction}, and four real-world SO-101 manipulation tasks~\cite{so101}.
LIBERO comprises 40 tasks across four suites (Spatial, Object, Goal, and LIBERO-10), with 50 rollouts per task.
For LIBERO-Plus, we report original and corrected-instruction results following La-JEPA~\cite{lajepa}, removing perturbation metadata from 8,493 non-language variants while preserving 1,537 language variants.
RoboTwin 2.0 follows the clean-to-randomized (C2R) protocol, while RoboTwin 2.0-Plus evaluates seven perturbations across 50 tasks with 10 paired rollouts per task.

\begin{figure*}[t]
    \centering
    \IfFileExists{figures/figure_real_frames_2col.pdf}{%
        \includegraphics[width=\linewidth]{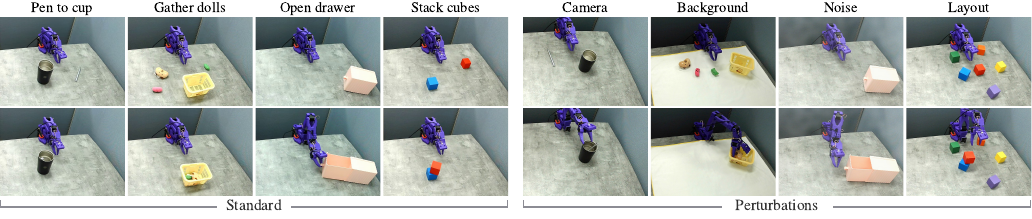}%
    }{%
        \fbox{\parbox[c][1.4in][c]{0.96\linewidth}{\centering
        REAL-ROBOT FRAME FIGURE PLACEHOLDER\\[4pt]
        Standard: Pen / Dolls / Drawer / Cube \quad $|$ \quad Perturbations: Camera / Background / Noise / Layout\\[4pt]
        Top: first frame, bottom: last frame
        }}%
    }
    \caption{\textbf{Real-world manipulation tasks and perturbations.}
Representative initial (top) and final (bottom) frames illustrating four manipulation tasks under standard conditions and four visual perturbation settings.}
    \label{fig:real_tasks}
\vspace{-2.0mm}
\end{figure*}

For real-world evaluation, we use four tasks: Pen to cup, Gather dolls, Open drawer, and Stack cubes (Fig.~\ref{fig:real_tasks}).
We collect 100 teleoperated demonstrations per task using LeRobot~\cite{cadene2026lerobot} and train a single multitask policy for 30k steps (global batch size 128).
The policy uses 640$\times$480 front and wrist RGB observations at 25 fps and a 6-dimensional joint state.
We evaluate 50 trials per task under standard conditions and 10 per task under each of four unseen perturbations (camera, background, sensor noise, and object layout), without further training.
Success requires stable stacking or pen placement after release, fully opening the drawer, or placing all three dolls in the basket.


\begin{table}[t]
\centering
\small
\caption{\textbf{RoboTwin 2.0 clean-to-randomized results.}
Success rates (\%). Emb. PT. and Extra Vid. PT. follow Table~\ref{tab:libero_corrected}.
Rand.: randomized environments; Avg.: mean of Clean and Rand.
$\dagger$: depth observations at inference.
Published results may differ in training and evaluation protocols and serve as reference comparisons.}
\label{tab:external}
\setlength{\tabcolsep}{5pt}
\resizebox{\columnwidth}{!}{%
\begin{tabular}{@{}lccccc@{}}
\toprule
\textbf{Method} &
\shortstack[c]{\textbf{Emb.}\\\textbf{PT.}} &
\shortstack[c]{\textbf{Extra}\\\textbf{Vid. PT.}} &
\textbf{Clean} &
\textbf{Rand.} &
\textbf{Avg.} \\
\midrule
\multicolumn{6}{@{}l}{\textcolor{gray}{\textit{VLA-based}}} \\ \midrule
$\pi_{0.5}$~\cite{pi05}
& $\checkmark$ & $\times$ & 70.7 & 46.0 & 58.4 \\
Qwen-RobotManip~\cite{qwenrobotmanip}
& $\checkmark$ & $\times$ & 84.7 & 69.4 & 77.1 \\
\midrule
\multicolumn{6}{@{}l}{\textcolor{gray}{\textit{WAM-based}}} \\ \midrule
Fast-WAM~\cite{fastwam}
& $\times$ & $\times$ & 77.8 & 1.9 & 39.9 \\
ImageWAM~\cite{imagewam}
& $\times$ & $\times$ & 85.0 & 37.6 & 61.3 \\
4D-WAM~\cite{4dwam}
& $\checkmark$ & $\times$ & 81.5 & 41.8 & 61.7 \\
OpenWAM-$\alpha$~\cite{openwam}
& $\checkmark$ & $\checkmark$ & 89.4 & 48.7 & 69.0 \\
PrismWAM$^\dagger$~\cite{prismwam}
& $\times$ & $\times$ & 88.0 & 53.2 & 70.6 \\
La-JEPA S~\cite{lajepa}
& $\times$ & $\times$ & 85.5 & 33.9 & 59.7 \\
NAVA-WAM~\cite{navawam}
& $\times$ & $\checkmark$ & 88.5 & 73.6 & 81.1 \\
ProWAM~\cite{prowam}
& $\times$ & $\checkmark$ & 85.8 & 75.7 & 80.8 \\
\midrule
\multicolumn{6}{@{}l}{\textcolor{gray}{\textit{Our implementation}}} \\ \midrule
Simple-WAM
& $\times$ & $\times$ & 80.2 & 2.7 & 41.5 \\
\rowcolor{oursbg}\textbf{Event-WAM}
& $\times$ & $\times$ & 80.6 & 1.9 & 41.3 \\
\bottomrule
\end{tabular}%
}
\end{table}

\begin{table}[t]
\centering
\small
\caption{\textbf{Zero-shot robustness on RoboTwin 2.0-Plus.}
Success rates (\%) over 50 tasks and 10 paired rollouts per task under
seven perturbations (7,000 rollouts across both methods).
Cam.: camera; Rob.: robot initial states; Bg.: background;
Lay.: object layouts; Avg.: trial-weighted mean.}
\label{tab:robotwin_plus}
\setlength{\tabcolsep}{2pt}
\resizebox{\columnwidth}{!}{%
\begin{tabular}{@{}lcccccccc@{}}
\toprule
\multirow{2}{*}{\textbf{Method}} &
\multicolumn{8}{c}{\textbf{RoboTwin 2.0-Plus}} \\
\cmidrule(lr){2-9}
& \textbf{Cam.} & \textbf{Rob.} & \textbf{Lang.} &
\textbf{Light} & \textbf{Bg.} & \textbf{Noise} &
\textbf{Lay.} & \textbf{Avg.} \\
\midrule
Simple-WAM
& 18.2 & 58.8 & 81.2 & \textbf{63.0} & \textbf{4.2}
& 53.7 & 25.8 & 43.6 \\
\rowcolor{oursbg}\textbf{Event-WAM}
& \textbf{19.6} & \textbf{62.0} & \textbf{81.8} & 62.8 & 3.8
& \textbf{60.5} & \textbf{32.2} & \textbf{46.1} \\
\bottomrule
\end{tabular}%
}
\vspace{-3mm}
\end{table}

\subsection{LIBERO and LIBERO-Plus}
\label{sec:exp_libero}

Table~\ref{tab:libero_corrected} compares Event-WAM with representative VLA and WAM methods under original and corrected LIBERO-Plus instructions. Under the corrected protocol, Event-WAM achieves 85.9\% success, improving upon our RGB-only Simple-WAM baseline (84.0\%) by 1.9 percentage points (pp). A consistent gain is observed under the original protocol (81.0\% vs. 78.9\%, +2.1 pp), while maintaining comparable performance on standard LIBERO (98.8\% vs. 98.9\%).
Event-WAM also achieves competitive robustness among recent approaches without additional embodied pretraining. Under corrected instructions, it exceeds the reported results of Fast-WAM (70.8\%), ImageWAM (80.4\%), and Cosmos Policy (84.9\%), while remaining close to NowWAM (86.2\%) and La-JEPA S (86.3\%). 
La-JEPA L achieves higher robustness (88.0\%). 
Since published methods differ in training and evaluation settings, these results serve as reference comparisons rather than matched rankings. 
Overall, Event-WAM improves zero-shot generalization over our RGB-only baseline while preserving strong in-distribution performance.

\subsection{RoboTwin 2.0 and RoboTwin 2.0-Plus}
\label{sec:exp_robotwin}

We further evaluate Event-WAM on RoboTwin 2.0 under the clean-to-randomized (C2R) protocol and on RoboTwin 2.0-Plus across seven perturbation types. As shown in Table~\ref{tab:external}, Event-WAM maintains comparable clean-environment performance to our RGB-only baseline (80.6\% vs. 80.2\%), while showing a slight decrease under randomization (1.9\% vs. 2.7\%).
In contrast, Table~\ref{tab:robotwin_plus} shows that Event-WAM improves average success from 43.6\% to 46.1\% (+2.5 pp), outperforming the baseline in five of seven perturbation categories, particularly under sensor noise (+6.8 pp) and object-layout changes (+6.4 pp).
These results suggest that the robustness benefits of event-derived supervision depend on the evaluation setting and perturbation type. Published methods differ in pretraining and evaluation protocols, so their results serve as reference comparisons rather than matched rankings.

\subsection{Real-World Manipulation}
\label{sec:exp_real}

We evaluate Event-WAM on physical robot manipulation tasks to assess whether its benefits extend beyond simulation.
Event-WAM and the RGB-only baseline share the same training data, budget, and RGB and proprioceptive inputs; neither requires an event camera at deployment.

\begin{figure}[t]
    \centering
    \includegraphics[width=\columnwidth]{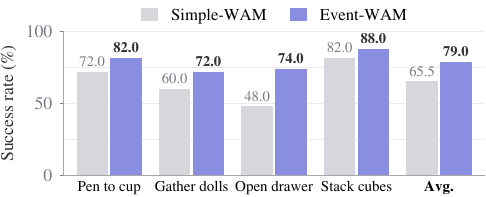}
    \vspace{-3mm}
   \caption{\textbf{Real-world performance.} 50 trials per task.}

    \label{fig:real_robot}
    \vspace{-1mm}
\end{figure}

\begin{figure}[t]
    \centering
    \includegraphics[width=\columnwidth]{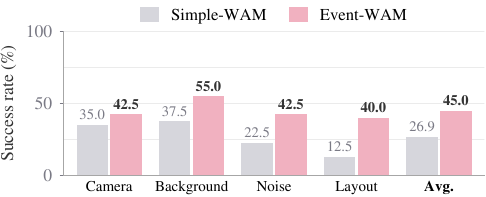}
    \vspace{-3mm}
\caption{\textbf{Real-world robustness.} 40 trials per condition.}
    \label{fig:real_robot_perturb}
    \vspace{-1mm}
\end{figure}

As shown in Fig.~\ref{fig:real_robot}, Event-WAM improves average success from 65.5\% to 79.0\% and outperforms Simple-WAM on all four tasks, with the largest gain on Open drawer (+26.0 pp).
Under zero-shot perturbations (Fig.~\ref{fig:real_robot_perturb}), Event-WAM increases average success from 26.9\% to 45.0\%, exceeding Simple-WAM across all four conditions, particularly under object-layout changes (+27.5 pp). 
These real-world results demonstrate the potential of event-derived future supervision to benefit RGB-based action learning under both standard and visually perturbed conditions.

\subsection{Ablation Study}
\label{sec:exp_ablations}

\begin{figure}[hbt!]
    \centering
    \includegraphics[width=\columnwidth]
    {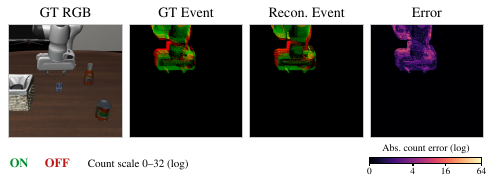}
    \caption{\textbf{Event reconstruction with a frozen RGB VAE.}
    Ground-truth and reconstructed event maps with errors.}
    \label{fig:event_vae_recon}
\end{figure}

\paragraph{Event reconstruction with a pretrained RGB VAE.}
Our encoding achieves 77.6\% event activity F1 and reduces active-region reconstruction error by 52\% over an all-zero baseline (Table~\ref{tab:event_vae_recon}), suggesting that a frozen RGB video VAE can represent event targets (Fig.~\ref{fig:event_vae_recon}).

\begin{table}[hbt!]
\centering
\small
\caption{\textbf{Event reconstruction using a frozen RGB VAE.}
Results on 32 policy rollouts (99 windows).
MAE is measured in raw event counts; Active MAE considers GT-active regions.}
\label{tab:event_vae_recon}
\setlength{\tabcolsep}{10pt}
\renewcommand{\arraystretch}{.5}
\resizebox{\columnwidth}{!}{%
\begin{tabular}{lccc}
\toprule
\textbf{Method} &
\textbf{F1 $\uparrow$} &
\textbf{Count MAE $\downarrow$} &
\textbf{Active MAE $\downarrow$} \\
\midrule
All-zero & 0.0 & 0.442 & 4.245 \\[1mm]
Wan VAE & \textbf{77.6} & \textbf{0.254} & \textbf{2.030} \\
\bottomrule
\end{tabular}%
}
\vspace{-2.0mm}
\end{table}

\paragraph{Joint RGB--Event loss weighting.}
We examine RGB and Event flow-matching weights while keeping the action loss fixed. As reported in Table~\ref{tab:joint_lambda}, all event-supervised variants numerically improve corrected LIBERO-Plus success over RGB-only (84.6--85.9\% vs. 84.0\%), while maintaining comparable LIBERO performance (98.7--99.0\%). Joint $(3,3)$ achieves the highest overall success (85.9\%) and leads under camera and layout perturbations, 
whereas Event-only supervision $(0,3)$ nearly matches its average (85.8\%) and performs best under background and sensor noise. 
These results suggest that event-derived future modeling can provide useful supervision for action learning even without an RGB future-prediction loss, with distinct robustness patterns across loss weightings.











\begin{table}[t]
\centering
\small
\caption{\textbf{Joint RGB--Event loss weighting on corrected LIBERO-Plus.}
Success rates (\%). Cam.: camera; Rob.: robot initial states;
Lang.: language; Light: lighting; Bg.: background;
Noise: sensor noise; Lay.: object layouts;
Avg.: weighted mean.}
\label{tab:joint_lambda}
\setlength{\tabcolsep}{2pt}
\resizebox{\columnwidth}{!}{%
\begin{tabular}{@{}ccccccccccc@{}}
\toprule
\multirow{2}{*}{$\boldsymbol{\lambda_{\mathrm{RGB}}}$} &
\multirow{2}{*}{$\boldsymbol{\lambda_{\mathrm{Event}}}$} &
\multirow{2}{*}{\shortstack[c]{\textbf{LIBERO}\\\textbf{Avg.}}} &
\multicolumn{8}{c}{\textbf{LIBERO-Plus}} \\
\cmidrule(lr){4-11}
& & & \textbf{Cam.} & \textbf{Rob.} & \textbf{Lang.} &
\textbf{Light} & \textbf{Bg.} & \textbf{Noise} &
\textbf{Lay.} & \textbf{Avg.} \\
\midrule
\rowcolor{baselinebg}
3 & 0 & 98.9 & 64.1 & \textbf{86.5} & 96.2 & \textbf{98.4} &
80.7 & 82.5 & 83.4 & 84.0 \\
\midrule
3 & 1 & 99.0 & 67.9 & 86.1 & 94.7 & 97.5 &
79.6 & 86.4 & 83.8 & 84.8 \\
1 & 3 & 98.8 & 69.7 & 82.5 & 91.3 & 97.1 &
82.2 & 88.7 & 83.8 & 84.6 \\
1 & 1 & 98.7 & 68.7 & 82.3 & \textbf{96.7} & 95.7 &
76.3 & 89.6 & 84.5 & 84.7 \\
0 & 3 & 98.9 & 69.2 & 82.8 & 94.5 & 95.7 &
\textbf{86.3} & \textbf{91.5} & 83.9 & 85.8 \\
\midrule
\rowcolor{oursbg}
3 & 3 & 98.8 & \textbf{70.9} & \textbf{86.5} & 94.5 & 95.4 &
81.2 & 88.9 & \textbf{85.6} & \textbf{85.9} \\
\bottomrule
\end{tabular}%
}
\vspace{-1.mm}
\end{table}

\begin{table}[hbt!]
\centering
\small
\caption{\textbf{Comparison of visual-change targets.}
Success rates (\%). Cam.: camera; Rob.: robot initial states; Bg.: background; Lay.: object layouts. Avg.: trial-count-weighted mean.}
\label{tab:diff_event}
\setlength{\tabcolsep}{1pt}
\resizebox{\columnwidth}{!}{%
\begin{tabular}{@{}lccccccccc@{}}
\toprule
& & \multicolumn{8}{c}{\textbf{LIBERO-Plus}} \\
\cmidrule(lr){3-10}
\textbf{Method} &
\textbf{LIBERO} &
\textbf{Cam.} &
\textbf{Rob.} &
\textbf{Lang.} &
\textbf{Light} &
\textbf{Bg.} &
\textbf{Noise} &
\textbf{Lay.} &
\textbf{Avg.} \\
\midrule
RGB-Diff
& 98.4 & \textbf{70.9} & 86.3 & 94.0 & \textbf{95.9}
& 78.5 & 87.0 & 84.9 & 85.2 \\[0.7mm]

\rowcolor{oursbg}
\textbf{Event-WAM}
& \textbf{98.8} & \textbf{70.9} & \textbf{86.5} & \textbf{94.5} & 95.4
& \textbf{81.2} & \textbf{88.9} & \textbf{85.6} & \textbf{85.9} \\
\bottomrule
\end{tabular}%
}
\end{table}

\paragraph{Visual-Change targets: RGB-Diff vs. Event.}
\label{sec:exp_diff_event}
Within the same WAM framework, we compare event-based future supervision with RGB-difference targets under otherwise matched settings. As shown in Table~\ref{tab:diff_event}, Event-WAM slightly outperforms RGB-Diff on LIBERO (98.8\% vs. 98.4\%) and corrected LIBERO-Plus (85.9\% vs. 85.2\%). Event supervision yields larger gains under background (+2.7 pp) and sensor noise (+1.9 pp), with a smaller improvement under object layout (+0.7 pp), while RGB-Diff performs slightly better under lighting (+0.5 pp). These differences suggest distinct robustness characteristics without establishing a specific causal mechanism.

\begin{table}[t]
\centering
\small
\caption{\textbf{Inference latency comparison.}
Latency (ms) on a single RTX A4000 GPU with BF16, batch size 1, and 10 action denoising steps. Measured over 80 runs per model.}
\label{tab:inference_latency}
\renewcommand{\arraystretch}{.9}
\setlength{\tabcolsep}{10pt}
\resizebox{.8\columnwidth}{!}{%
\begin{tabular}{lcc}
\toprule
\textbf{Method} &
\textbf{Mean (ms) $\downarrow$} &
\textbf{P95 (ms) $\downarrow$} \\
\midrule
Simple-WAM & 531.24 & 534.38 \\[0.5mm]
\textbf{Event-WAM} & 534.11 & 539.00 \\
\bottomrule
\end{tabular}}
\vspace{-3.0mm}
\end{table}

\begin{figure*}[t]
    \centering 
    \includegraphics[width=\textwidth]
    {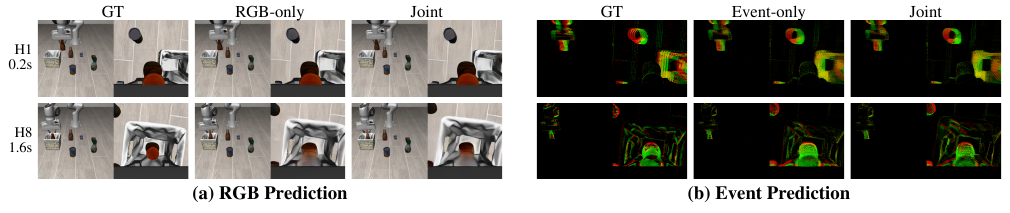}

\caption{\textbf{Qualitative future prediction comparison on LIBERO.}
(a) RGB predictions from RGB-only and Joint models.
(b) Event predictions from Event-only and Joint models.
H1 and H8 denote the nearest (0.2\,s) and farthest (1.6\,s) steps, respectively. Predictions use NFE=1. Head and wrist views are concatenated side by side.
Event visualizations distinguish ON (green) and OFF (red) polarities.}

    \label{fig:prediction_quality}
    \vspace{-2.mm}
\end{figure*}

\paragraph{Inference efficiency.}
Event-WAM incurs only 2.87\,ms (0.54\%) additional mean inference latency compared with RGB-only Simple-WAM (Table~\ref{tab:inference_latency}). Measurements include input preprocessing and action chunk generation, excluding simulator execution, communication, and model loading.




\subsection{Analysis}
\label{sec:prediction_quality}

\paragraph{Future prediction quality.}
We evaluate future prediction quality on 99 windows from 32 recorded policy trajectories, using three shared noise realizations per window and both head and wrist views. We examine the nearest (H1, 0.2\,s) and farthest (H8, 1.6\,s) future steps with NFE=1 and NFE=10, where NFE denotes video flow-velocity evaluations. Event targets are derived from the corresponding RGB streams.
As summarized in Table~\ref{tab:prediction_quality}, single-target and Joint models achieve comparable prediction quality, with all paired 95\% confidence intervals including zero. Prediction quality declines substantially from H1 to H8, while increasing NFE to 10 provides no consistent improvement. Figure~\ref{fig:prediction_quality} provides qualitative comparisons at both horizons. Overall, the policy-level benefits of joint supervision are not consistently accompanied by improved future prediction fidelity.
For single-pass WAMs that condition actions on future-token features rather than reconstructed frames, this suggests that prediction fidelity alone may not fully capture the value of future supervision for action learning.

\begin{figure}[t]
    \centering
    \includegraphics[width=\linewidth]{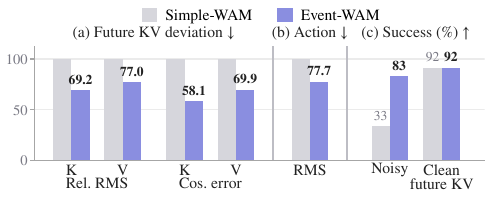}
\vspace{-4.0mm}
    \caption{\textbf{Robustness to image corruption.}
    (a) Relative RMS, $\|\tilde{X}-X\|_F/\|X\|_F$, and cosine error, $1-\cos(\tilde{X},X)$, of future keys (K) and values (V), where $X$ and $\tilde{X}$ denote each model's clean and corrupted features.
(b) RMS of noise-induced action changes through future KV over the first 10 executed actions.
(a,b) are normalized to Simple-WAM${}=100$.
(c) Success rates (\%) over 12 common states (six sensor-noise cases, two seeds), without (Noisy) and with clean future KV.
$\downarrow$/$\uparrow$: lower/higher is better.
    }
    \label{fig:robustness_analysis}
\vspace{-2.0mm}
\end{figure}

\paragraph{Robustness to image corruption.}
We further examine the robustness of future-token features on 60 observations from 30 tasks, applying five types of image noise to the head camera while keeping all other inputs fixed. Deviations are measured against each model's clean forward pass.
As shown in Fig.~\ref{fig:robustness_analysis}(a), Event-WAM reduces the relative RMS of future keys and values by 30.8\% and 23.0\% compared with Simple-WAM, and their cosine error by 41.9\% and 30.1\%, with lower values in all 30 tasks.
Correspondingly, over the first 10 executed actions, the noise-induced action change through future KV decreases by 22.3\% (corrected $p<0.001$; Fig.~\ref{fig:robustness_analysis}(b)).
On six sensor-noise cases, Event-WAM achieves 83\% success, compared with 33\% for Simple-WAM.
Replacing the future KV with clean values raises Simple-WAM to 92\%, whereas replacing only the current KV yields 58\% (Fig.~\ref{fig:robustness_analysis}(c)).
These results indicate that event supervision makes future-token features more robust to image corruption, in line with the 6.4\,pp gain under sensor noise in Table~\ref{tab:joint_lambda}.

\begin{table}[t]
\centering
\small
\caption{\textbf{Future prediction quality on LIBERO.}
H1 and H8 correspond to 0.2\,s and 1.6\,s futures.
NFE denotes the number of video flow-velocity evaluations.
Only refers to RGB-only for RGB prediction and Event-only for event prediction.}
\label{tab:prediction_quality}
\setlength{\tabcolsep}{6pt}
\renewcommand{\arraystretch}{.96}
\resizebox{\columnwidth}{!}{%
\begin{tabular}{@{}llccccc@{}}
\toprule
\multirow{2}{*}{\textbf{Target}} &
\multirow{2}{*}{\textbf{Metric}} &
\multirow{2}{*}{\textbf{NFE}} &
\multicolumn{2}{c}{\textbf{H1 (0.2\,s)}} &
\multicolumn{2}{c}{\textbf{H8 (1.6\,s)}} \\
\cmidrule(lr){4-5}
\cmidrule(lr){6-7}
& & & \textbf{Only} & \textbf{Joint} &
\textbf{Only} & \textbf{Joint} \\
\midrule
\multirow{4}{*}{RGB}
& \multirow{2}{*}{PSNR $\uparrow$ (dB)}
& 1  & 27.02 & 26.88 & 21.26 & 21.26 \\
& & 10 & 26.91 & 26.86 & 21.01 & 21.08 \\
& \multirow{2}{*}{SSIM $\uparrow$}
& 1  & 0.867 & 0.864 & 0.780 & 0.781 \\
& & 10 & 0.865 & 0.864 & 0.776 & 0.778 \\
\midrule
\multirow{2}{*}{Event}
& \multirow{2}{*}{Activity F1 $\uparrow$ (\%)}
& 1  & 78.1 & 77.9 & 64.0 & 63.9 \\
& & 10 & 78.0 & 77.8 & 63.7 & 63.4 \\
\bottomrule
\end{tabular}%
}
\vspace{-2.mm}
\end{table}

\section{Conclusion}
\label{sec:conclusion}
We presented Event-WAM, which incorporates event-derived visual changes as complementary future supervision alongside RGB for robotic action learning. By jointly modeling RGB and event-derived futures, Event-WAM provides appearance- and change-oriented supervision without requiring event observations or explicit future generation at inference. Across simulated and real-world manipulation benchmarks, Event-WAM maintains comparable performance under standard simulated conditions while demonstrating environment-dependent robustness gains and substantial improvements in real-world manipulation. Further analyses show that event-only supervision approaches joint supervision performance and that policy improvements are not consistently accompanied by higher future prediction fidelity. 
These findings suggest that event-derived supervision may benefit action learning not by introducing new sensory information or improving prediction accuracy, but by explicitly emphasizing action-relevant visual changes, providing a useful inductive bias for robotic manipulation.

\paragraph{Limitations.}
Our event representations are synthesized from RGB rather than captured by event cameras. Moreover, temporal binning preserves coarse temporal structure but discards fine-grained event ordering within each bin, while the frozen RGB video VAE is not designed for event dynamics. Future work could explore real event-camera data and event-aware representations that exploit fine-grained temporal information for action learning.

\clearpage


{
    \small
    \bibliographystyle{ieeenat_fullname}
    \bibliography{main}
}



\clearpage
\setcounter{page}{1}
\maketitlesupplementary
\appendix
\raggedbottom

\setcounter{table}{0}
\setcounter{figure}{0}
\setcounter{equation}{0}
\renewcommand{\thetable}{S\arabic{table}}
\renewcommand{\thefigure}{S\arabic{figure}}
\renewcommand{\theequation}{S\arabic{equation}}

\section{Implementation Details}
\label{sec:supp_impl}

We build Event-WAM upon Simple-WAM~\cite{zhou2026makes}, initializing the video model from Wan2.2-TI2V-5B~\cite{wan2025} and following the Simple-WAM action model, first-frame-causal video attention, and language and state conditioning.
For LIBERO~\cite{libero}, each training window contains 33 RGB frames at 20\,Hz, subsampled at a stride of 4 to obtain 9 frames for encoding.
LIBERO uses a head and a wrist camera, whereas RoboTwin 2.0~\cite{robotwin2} uses one head and two wrist cameras.
For LIBERO, both views are resized to $224\times224$ and concatenated horizontally, and the action chunk spans 32 control steps.
Events synthesized using v2e~\cite{v2e} are accumulated into 8 temporal bins of 0.2\,s each, aligned with the RGB subsampling intervals (Sec.~\ref{sec:supp_event}).
For event log-count normalization, we use $\gamma_E=3.4965$ for LIBERO and $3.0910$ for RoboTwin 2.0.
Table~\ref{tab:supp_hparams} summarizes the training hyperparameters.
Unless otherwise specified, all models use the same training data, optimization settings, and training budget.
At inference, the action model predicts a 32-step action chunk with 10 flow-matching steps.
For RoboTwin 2.0, we use 2,500 clean demonstrations (50 tasks, 50 per task) under the clean-to-randomized protocol and train for 21,125 steps using the same optimizer and loss weights, replanning every 24 control steps.

\begin{table}[!h]
\centering
\caption{\textbf{Training hyperparameters on LIBERO.}
Loss weights correspond to Event-WAM; variant-specific changes are described in the respective experiments.}
\label{tab:supp_hparams}
\setlength{\tabcolsep}{20pt}
\resizebox{\columnwidth}{!}{%
\begin{tabular}{@{}ll@{}}
\toprule
Setting & Value \\
\midrule
Optimizer & AdamW ($\beta_1{=}0.9$, $\beta_2{=}0.95$) \\
Learning rate & $1\times10^{-4}$, cosine, 5\% warmup \\
Weight decay & 0.01 \\
Training steps & 21{,}700 \\
Global batch size & 128 \\
Precision & BF16 \\
$\lambda_{\mathrm{RGB}}$ / $\lambda_{\mathrm{Event}}$ / $\lambda_{\mathrm{act}}$ & 3 / 3 / 1 \\
$p_1$ / noise schedule shift & 0.5 / 5 \\
Hardware & $4\times$ A100 80GB ($\sim$29\,h) \\
\bottomrule
\end{tabular}}
\end{table}

\FloatBarrier
\section{Event Synthesis and Representation}
\label{sec:supp_event}

\subsection{Event Simulation and Temporal Binning}
\label{sec:supp_event_sim}
We synthesize events offline from the RGB demonstration videos using the v2e simulator.
Each camera stream is resized, converted to grayscale, and simulated independently with ON and OFF contrast thresholds of 0.2 and a per-pixel threshold mismatch of $\sigma=0.03$.
We disable leak events, shot noise, and the photoreceptor low-pass filter, and process frames at their native rate without interpolation.
These settings are shared across all datasets (Table~\ref{tab:supp_v2e}).

The generated events are accumulated into polarity-separated per-pixel counts (Eq.~(4) of the main paper).
Each temporal bin covers four consecutive RGB transitions, matching the stride-4 RGB subsampling, so the eight bins span the 32 transitions between the observation frame and the last future frame; the bin duration therefore follows the frame rate of each dataset.
Transitions beyond the episode end contribute zero counts, and future latents lying entirely beyond it are excluded from both the RGB and event losses.

\begin{table}[!h]
\centering
\caption{\textbf{Event synthesis settings.} Settings in the bottom block are shared by all datasets.
$\gamma_E$ is estimated per dataset.}
\label{tab:supp_v2e}
\setlength{\tabcolsep}{3pt}
\resizebox{\columnwidth}{!}{%
\begin{tabular}{@{}lccc@{}}
\toprule
 & LIBERO & RoboTwin 2.0 & Real Robot \\
\midrule
Episodes            & 1{,}712         & 2{,}500                    & 400 \\
RGB frame rate      & 20\,Hz          & 50\,Hz                     & 25\,Hz \\
Cameras             & 2               & 3                          & 2 \\
Event synthesis resolution & $224{\times}224$ & \begin{tabular}[c]{@{}c@{}}Head: $256{\times}320$\\Wrist: $128{\times}160$\end{tabular} & $224{\times}224$ \\
Bin duration        & 0.2\,s          & 0.08\,s                    & 0.16\,s \\
$\gamma_E$          & 3.4965          & 3.0910                     & 3.2189 \\
\midrule
Transitions per bin / bins   & \multicolumn{3}{c}{4 / 8} \\
Contrast threshold (ON/OFF)  & \multicolumn{3}{c}{0.2 / 0.2} \\
Threshold mismatch $\sigma$  & \multicolumn{3}{c}{0.03} \\
Leak / shot noise / low-pass & \multicolumn{3}{c}{disabled} \\
Frame interpolation          & \multicolumn{3}{c}{none} \\
\bottomrule
\end{tabular}}
\end{table}

\subsection{Pseudo-RGB Encoding}
\label{sec:supp_event_enc}
\begin{figure*}[!t]
\centering
\includegraphics[width=\textwidth]{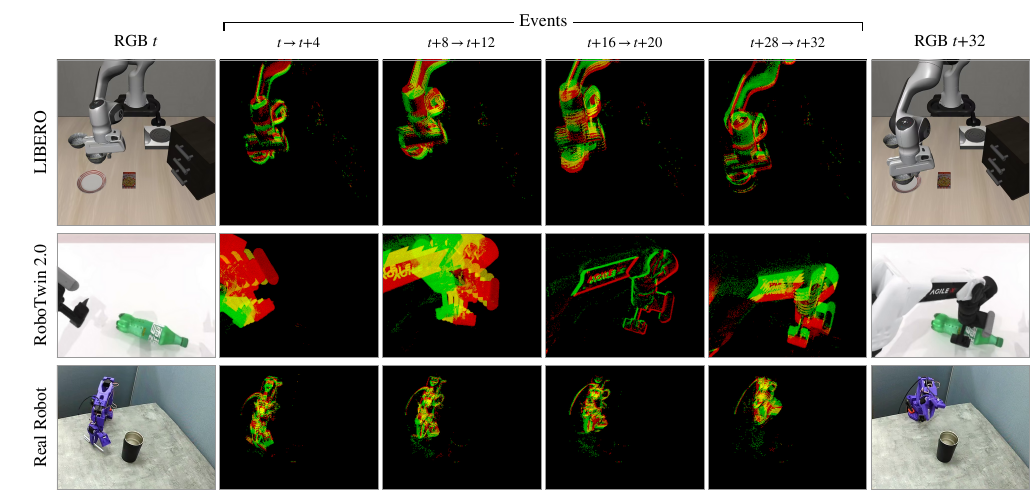}
\caption{\textbf{Synthesized events across datasets.} For each dataset, we show the observation frame, four of the eight event bins, and the last future frame of a training window (one camera view). ON and OFF events are shown in green and red, respectively. Events concentrate on the moving robot and manipulated objects, while the static background produces few events.}
\label{fig:supp_event_examples}
\end{figure*}

The polarity-separated counts are log-transformed and mapped to three-channel pseudo-RGB frames following Eq.~(6) of the manuscript, and the result is clipped to the VAE input range $[-1,1]$.
For each dataset, the normalization scale $\gamma_E$ is
the 99.9th percentile of $\log(1+n)$ over all nonzero
counts, pooled across both polarities, all eight bins,
and all camera views of randomly sampled training
windows, limiting saturation of active event values.
An event-free pixel maps to $-1$ in all channels, so the reference frame $E_0$ (Eq.~(7)) is the representation of a scene with no brightness change.
The pseudo-RGB event frames are concatenated across camera views in the same spatial layout as the RGB frames and encoded by the frozen Wan2.2 VAE into event latents that are spatially and temporally aligned with the RGB future latents.
Figure~\ref{fig:supp_event_examples} shows representative event sequences from each dataset.

\section{Extended Quantitative Results}
\label{sec:supp_results}

\subsection{Per-Suite Results on LIBERO}
\label{sec:supp_libero}

Table~\ref{tab:supp_libero_per_suite} reports per-suite success rates on LIBERO.
Event-WAM achieves comparable overall performance to Simple-WAM (98.8\% vs.\ 98.9\%), with modest suite-dependent differences.
Event-WAM improves Spatial success by 1.2 percentage points (pp), while showing a 1.4 pp decrease on LIBERO-10. 
Object performance is identical, and Goal performance differs by only 0.4 pp.

\begin{table}[!h]
\centering
\caption{\textbf{LIBERO per-suite performance.} Success rates (\%).} 
\label{tab:supp_libero_per_suite}
\scriptsize
\setlength{\tabcolsep}{3pt}
\resizebox{\columnwidth}{!}{%
\begin{tabular}{lccccc}
\toprule
\textbf{Method} & \textbf{Spatial} & \textbf{Object} & \textbf{Goal} & \textbf{LIBERO-10} & \textbf{Overall} \\
\midrule
Simple-WAM              & 98.2 & 100.0 & 98.8 & 98.6 & 98.9 \\
\rowcolor{oursbg}Event-WAM       & 99.4 & 100.0 & 98.4 & 97.2 & 98.8 \\
\bottomrule
\end{tabular}}
\end{table}

\subsection{Per-Task Results on RoboTwin 2.0}
\label{sec:supp_robotwin}
Table~\ref{tab:supp_robotwin_per_task} reports per-task success rates under clean and randomized conditions. 
Under clean conditions, Event-WAM performs comparably to Simple-WAM (80.60\% vs.\ 80.24\%), with varying results across tasks.
Under randomization, both models degrade sharply, with success rates of at most 3\% on 43 of the 50 tasks (1.94\% vs.\ 2.72\% overall).

Notably, the overall difference is entirely attributable to \texttt{click\_alarmclock} (20\% vs.\ 59\%), while differences
across the remaining 49 tasks cancel out in aggregate.
Further analysis reveals a larger gap in right-arm episodes (5/55 vs.\ 33/55) than in left-arm episodes (15/45 vs.\ 26/45), warranting further investigation into the underlying causes.
Despite lower overall success, Event-WAM achieves at least one success on more randomized tasks (25/50 vs.\ 17/50).
However, the 11 tasks uniquely solved by Event-WAM have only one or two successes each, indicating broader observed task coverage rather than reliable execution.


\begin{table*}[!t]
\centering
\caption{\textbf{RoboTwin 2.0 clean-to-randomized performance.}
Success rates (\%) across 50 tasks, with 100 trials per task per condition.}
\label{tab:supp_robotwin_per_task}
\fontsize{10.pt}{11.pt}\selectfont
\begin{tabular*}{\textwidth}{@{\extracolsep{\fill}}lcccc@{}}
\toprule
& \multicolumn{2}{c}{\textbf{Clean}} & \multicolumn{2}{c}{\textbf{Randomized}} \\
\cmidrule(lr){2-3}\cmidrule(l){4-5}
\textbf{Task} & \textbf{Simple-WAM} & \textbf{Event-WAM} & \textbf{Simple-WAM} & \textbf{Event-WAM} \\
\midrule
\texttt{adjust\_bottle} & 98 & 100 & 0 & 2 \\
\texttt{beat\_block\_hammer} & 92 & 90 & 0 & 2 \\
\texttt{blocks\_ranking\_rgb} & 96 & 98 & 1 & 1 \\
\texttt{blocks\_ranking\_size} & 43 & 48 & 0 & 0 \\
\texttt{click\_alarmclock} & 98 & 100 & 59 & 20 \\
\texttt{click\_bell} & 98 & 100 & 10 & 7 \\
\texttt{dump\_bin\_bigbin} & 93 & 94 & 1 & 0 \\
\texttt{grab\_roller} & 100 & 100 & 3 & 2 \\
\texttt{handover\_block} & 82 & 83 & 0 & 0 \\
\texttt{handover\_mic} & 98 & 98 & 0 & 0 \\
\texttt{hanging\_mug} & 35 & 39 & 0 & 0 \\
\texttt{lift\_pot} & 93 & 98 & 0 & 0 \\
\texttt{move\_can\_pot} & 90 & 85 & 0 & 2 \\
\texttt{move\_pillbottle\_pad} & 91 & 89 & 0 & 0 \\
\texttt{move\_playingcard\_away} & 95 & 96 & 0 & 1 \\
\texttt{move\_stapler\_pad} & 18 & 17 & 0 & 0 \\
\texttt{open\_laptop} & 91 & 89 & 1 & 0 \\
\texttt{open\_microwave} & 38 & 31 & 0 & 1 \\
\texttt{pick\_diverse\_bottles} & 60 & 67 & 1 & 0 \\
\texttt{pick\_dual\_bottles} & 77 & 73 & 8 & 5 \\
\texttt{place\_a2b\_left} & 79 & 77 & 1 & 1 \\
\texttt{place\_a2b\_right} & 77 & 77 & 0 & 0 \\
\texttt{place\_bread\_basket} & 92 & 91 & 2 & 4 \\
\texttt{place\_bread\_skillet} & 88 & 86 & 0 & 0 \\
\texttt{place\_burger\_fries} & 97 & 97 & 2 & 2 \\
\texttt{place\_can\_basket} & 69 & 76 & 0 & 0 \\
\texttt{place\_cans\_plasticbox} & 96 & 95 & 0 & 1 \\
\texttt{place\_container\_plate} & 96 & 97 & 1 & 2 \\
\texttt{place\_dual\_shoes} & 63 & 66 & 0 & 0 \\
\texttt{place\_empty\_cup} & 99 & 98 & 0 & 0 \\
\texttt{place\_fan} & 52 & 58 & 0 & 0 \\
\texttt{place\_mouse\_pad} & 63 & 56 & 0 & 0 \\
\texttt{place\_object\_basket} & 80 & 83 & 0 & 0 \\
\texttt{place\_object\_scale} & 85 & 84 & 0 & 0 \\
\texttt{place\_object\_stand} & 98 & 97 & 0 & 1 \\
\texttt{place\_phone\_stand} & 93 & 94 & 0 & 0 \\
\texttt{place\_shoe} & 71 & 73 & 0 & 1 \\
\texttt{press\_stapler} & 79 & 86 & 3 & 3 \\
\texttt{put\_bottles\_dustbin} & 70 & 69 & 0 & 2 \\
\texttt{put\_object\_cabinet} & 52 & 53 & 0 & 0 \\
\texttt{rotate\_qrcode} & 81 & 71 & 0 & 0 \\
\texttt{scan\_object} & 76 & 75 & 0 & 0 \\
\texttt{shake\_bottle\_horizontally} & 100 & 100 & 18 & 12 \\
\texttt{shake\_bottle} & 100 & 100 & 21 & 14 \\
\texttt{stack\_blocks\_three} & 90 & 92 & 0 & 0 \\
\texttt{stack\_blocks\_two} & 98 & 98 & 0 & 1 \\
\texttt{stack\_bowls\_three} & 66 & 82 & 0 & 2 \\
\texttt{stack\_bowls\_two} & 94 & 95 & 1 & 3 \\
\texttt{stamp\_seal} & 66 & 60 & 0 & 0 \\
\texttt{turn\_switch} & 56 & 49 & 3 & 5 \\
\midrule
\textbf{Overall (\%)} & \textbf{80.24} & \textbf{80.60} & \textbf{2.72} & \textbf{1.94} \\
\bottomrule
\end{tabular*}
\end{table*}

\subsection{Additional Ablation Studies}
\label{sec:supp_ablation}
\paragraph{Capacity-matched control.}
Event-WAM expands the video model's input and output projections from $C$ to $2C$ channels to jointly model RGB and event latents. 
To examine the effect of increased model capacity, we evaluate a capacity-matched Simple-WAM variant that retains the expanded RGB--auxiliary projections with zero-valued auxiliary targets, while disabling auxiliary future supervision.
As shown in Table~\ref{tab:supp_capacity}, the capacity-matched variant achieves comparable performance to Simple-WAM on LIBERO (98.7\% vs.\ 98.9\%) and identical overall success on LIBERO-Plus (84.0\%).
In contrast, Event-WAM achieves 85.9\% on LIBERO-Plus, with a particularly large improvement under sensor noise (88.9\% vs.\ 81.4\%).
These results suggest that the observed gains cannot be explained solely by the expanded model capacity.

\begin{table}[!h]
\centering
\caption{\textbf{Capacity-matched comparison.}
Success rates (\%) on LIBERO and corrected LIBERO-Plus.
$\dagger$: capacity-matched Simple-WAM with expanded
projections and no auxiliary future supervision.
Cam.: camera; Rob.: robot initial states; Lang.: language; Light: lighting; Bg.: background; Noise: sensor noise; Lay.: object layouts; Avg.: trial-weighted mean.}
\label{tab:supp_capacity}
\setlength{\tabcolsep}{1pt}
\resizebox{\columnwidth}{!}{%
\begin{tabular}{lccccccccc}
\toprule
& \textbf{LIBERO} & \multicolumn{8}{c}{\textbf{LIBERO-Plus}} \\
\cmidrule(lr){2-2}
\cmidrule(l){3-10}
\textbf{Method} & \textbf{Avg.} & \textbf{Cam.} & \textbf{Rob.} & \textbf{Lang.} & \textbf{Light}
& \textbf{Bg.} & \textbf{Noise} & \textbf{Lay.} & \textbf{Avg.} \\
\midrule
Simple-WAM
& \textbf{98.9} & 64.1 & \textbf{86.5} & \textbf{96.2}
& \textbf{98.4} & 80.7 & 82.5 & 83.4 & 84.0 \\
Simple-WAM$^\dagger$
& 98.7 & 67.1 & 86.3 & 95.4
& 98.0 & 79.5 & 81.4 & 83.3 & 84.0 \\
\rowcolor{oursbg} Event-WAM
& 98.8 & \textbf{70.9} & \textbf{86.5} & 94.5
& 95.4 & \textbf{81.2} & \textbf{88.9}
& \textbf{85.6} & \textbf{85.9} \\
\bottomrule
\end{tabular}}
\end{table}

\paragraph{Comparison of visual-change targets.}
We further compare event-derived supervision with RGB differences and optical flow under the same WAM framework.
Following DreamWAM~\cite{dreamwam}, we generate optical-flow targets using a pretrained RAFT~\cite{teed2020raft} model and encode them with the frozen RGB video VAE.
As shown in Table~\ref{tab:supp_flow_comparison}, Event-WAM achieves the highest overall success on LIBERO-Plus (85.9\%), compared with RGB-Diff (85.2\%) and Optical Flow (85.4\%).
While Optical Flow performs better under camera and sensor-noise perturbations, Event-WAM shows stronger robustness under robot initial-state and background changes, indicating distinct robustness patterns across visual-change targets.

\begin{table}[t]
\centering
\caption{\textbf{Comparison of visual-change targets.}
Success rates (\%) on LIBERO and corrected LIBERO-Plus.
Cam.: camera; Rob.: robot initial states; Lang.: language; Light: lighting; Bg.: background; Noise: sensor noise; Lay.: object layouts; Avg.: trial-weighted mean.}
\label{tab:supp_flow_comparison}
\scriptsize
\setlength{\tabcolsep}{1.pt}
\resizebox{\columnwidth}{!}{%
\begin{tabular}{lccccccccc}
\toprule
& \textbf{LIBERO} & \multicolumn{8}{c}{\textbf{LIBERO-Plus}} \\
\cmidrule(lr){2-2}
\cmidrule(l){3-10}
\textbf{Method} & \textbf{Avg.} & \textbf{Cam.} & \textbf{Rob.} & \textbf{Lang.} & \textbf{Light}
& \textbf{Bg.} & \textbf{Noise} & \textbf{Lay.} & \textbf{Avg.} \\
\midrule
RGB-Diff
& 98.4 & 70.9 & 86.3 & 94.0 & \textbf{95.9}
& 78.5 & 87.0 & 84.9 & 85.2 \\
Optical Flow
& 98.5 & \textbf{74.5} & 81.9 & 94.2 & 95.4
& 76.0 & \textbf{90.3} & \textbf{85.7} & 85.4 \\
\rowcolor{oursbg}Event-WAM
& \textbf{98.8} & 70.9 & \textbf{86.5}
& \textbf{94.5} & 95.4 & \textbf{81.2}
& 88.9 & 85.6 & \textbf{85.9} \\
\bottomrule
\end{tabular}%
}
\end{table}

\begin{table}[t]
\centering
\caption{\textbf{Event supervision across WAM architectures.}
Success rates (\%) on LIBERO and corrected LIBERO-Plus.
Cam.: camera; Rob.: robot initial states; Lang.: language; Light: lighting; Bg.: background; Noise: sensor noise; Lay.: object layouts; Avg.: trial-weighted mean.}
\label{tab:supp_backbone_generalization}
\scriptsize
\setlength{\tabcolsep}{1.pt}
\resizebox{\columnwidth}{!}{%
\begin{tabular}{lccccccccc}
\toprule
& \textbf{LIBERO} & \multicolumn{8}{c}{\textbf{LIBERO-Plus}} \\
\cmidrule(lr){2-2}
\cmidrule(l){3-10}
\textbf{Method} & \textbf{Avg.} & \textbf{Cam.} & \textbf{Rob.} & \textbf{Lang.} & \textbf{Light}
& \textbf{Bg.} & \textbf{Noise} & \textbf{Lay.} & \textbf{Avg.} \\
\midrule
Fast-WAM
& 97.1 & 51.5 & 73.2 & 71.6 & \textbf{93.3}
& 72.2 & \textbf{82.8} & 78.6 & 74.0 \\
\rowcolor{oursbg}{\quad + Event}
& \textbf{97.7} & \textbf{56.1} & \textbf{77.9} & \textbf{80.2} & 92.1
& \textbf{73.0} & 81.6 & \textbf{79.3} & \textbf{76.7} \\
\midrule
Faster-WAM
& 98.9 & \textbf{61.5} & 79.6 & 91.0 & 94.7
& 65.6 & 76.8 & 82.8 & 78.7 \\
\rowcolor{oursbg}{\quad + Event}
& \textbf{98.9} & 58.7 & \textbf{81.5} & \textbf{94.7} & \textbf{96.2}
& \textbf{71.8} & \textbf{81.8} & \textbf{84.2} & \textbf{81.0} \\
\midrule
Simple-WAM
& \textbf{98.9} & 64.1 & \textbf{86.5} & \textbf{96.2} & \textbf{98.4}
& 80.7 & 82.5 & 83.4 & 84.0 \\
\rowcolor{oursbg}{\quad + Event}
& 98.8 & \textbf{70.9} & \textbf{86.5} & 94.5 & 95.4
& \textbf{81.2} & \textbf{88.9} & \textbf{85.6} & \textbf{85.9} \\
\bottomrule
\end{tabular}%
}
\end{table}

\paragraph{Generalization across WAM architectures.}
We further examine the applicability of event-derived future supervision across Fast-WAM~\cite{fastwam}, Faster-WAM~\cite{fasterwam_future}, and Simple-WAM.
For all three architectures, we use the same event representation and joint RGB--event supervision objective.
We preserve each backbone's original architecture and conditioning scheme, expanding only the video input and output projections to accommodate event latents.
As shown in Table~\ref{tab:supp_backbone_generalization}, incorporating event supervision improves corrected LIBERO-Plus success from 74.0\% to 76.7\% on Fast-WAM, 78.7\% to 81.0\% on Faster-WAM, and 84.0\% to 85.9\% on Simple-WAM.
Notably, Fast-WAM also improves on standard LIBERO (97.1\% to 97.7\%), while the other architectures maintain comparable performance.
Although improvements vary across perturbation types, the consistent overall gains suggest that event-derived future supervision can benefit different WAM architectures.

\section{Alternative Event-WAM Formulation}
\label{sec:supp_variant}

\subsection{Residual-Aware Formulation}
\label{sec:supp_residual}

\begin{figure*}[t]
\centering
\includegraphics[width=\textwidth]{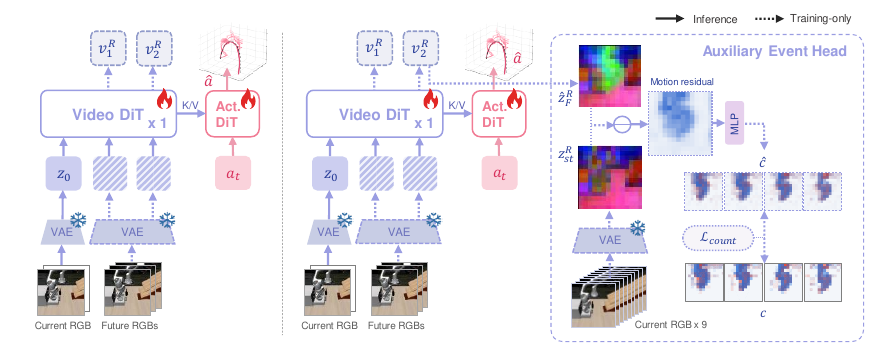}
\caption{\textbf{Simple-WAM (left) vs.\ residual-aware event supervision (right).}
The auxiliary head $g_\phi$ predicts event log-counts $\hat{\mathbf c}$ from the motion residual $\hat{\mathbf r}=\hat{\mathbf z}^R_F-\mathbf z^R_{st}$, and its loss $\mathcal L_{\mathrm{count}}$ is backpropagated into the video DiT.
Dotted paths are training-only, so inference is identical for both models.}
\label{fig:supp_residual}
\end{figure*}

Motivated by the change-oriented nature of events, we explore residual-aware event supervision as an alternative to joint RGB--event modeling.
This variant supervises predicted RGB latent transitions using event-derived targets while preserving the RGB-only generative pathway (Fig.~\ref{fig:supp_residual}).
From the predicted RGB velocity, we estimate the clean future latents
$\hat{\mathbf z}^R_F=\mathbf z^R_\sigma-\sigma\mathbf v^R_\theta$
and subtract a static reference $\mathbf z^R_{st}$, obtained by encoding the observation frame repeated across the window with the same VAE.

Unlike joint modeling, which encodes event sequences into VAE latents, this formulation directly supervises coarse polarity-separated event counts at the RGB latent spatial resolution.
The counts are organized into a configurable number of temporal bins per future latent group, allowing temporal granularity to vary without modifying the RGB generative pathway (Table~\ref{tab:supp_res_bin}).
A lightweight head $g_\phi$ (per-token LayerNorm and a two-layer MLP over $2\times2$ latent patches; 58k parameters with four bins and 66k with eight) maps the predicted transition
$\hat{\mathbf r}=\hat{\mathbf z}^R_F-\mathbf z^R_{st}$
to the event log-count targets.
Event-loss gradients propagate through the predicted RGB transition into the video model.

The training objective is
\begin{equation}
\label{eq:supp_res}
\mathcal L_{\mathrm{res}}
=\mathcal L_{\mathrm{act}}
+\lambda_{\mathrm{RGB}}\mathcal L_{\mathrm{RGB}}
+\lambda_{\mathrm{count}}\mathcal L_{\mathrm{count}},
\end{equation}
where $\mathcal L_{\mathrm{count}}$ is an MSE with doubled weight at event-active locations (where ground-truth event counts are nonzero) and $\lambda_{\mathrm{RGB}}=3$.
The auxiliary head and static reference are discarded at inference, leaving the Simple-WAM inference procedure unchanged.

\subsection{Experimental Results}
\label{sec:supp_residual_results}

Tables~\ref{tab:supp_res_lambda} and~\ref{tab:supp_res_bin} report the effects of event loss weight and temporal resolution on LIBERO and corrected LIBERO-Plus, against an RGB-only model trained under the same settings.
All variants maintain near-saturated LIBERO performance, with only modest changes on corrected LIBERO-Plus.
Table~\ref{tab:supp_res_quality} additionally reports RGB and event prediction quality across event loss weights.

\begin{table}[t]
\centering
\caption{\textbf{Event loss weight for residual-aware supervision.}
Success rates (\%) on LIBERO and corrected LIBERO-Plus.
Residual variants use 4 temporal bins (0.2\,s each) per future latent group.
Cam.: camera; Rob.: robot initial states; Lang.: language; Light: lighting; Bg.: background; Noise: sensor noise; Lay.: object layouts; Avg.: trial-weighted mean.
Bold: best per column.}
\label{tab:supp_res_lambda}
\footnotesize
\setlength{\tabcolsep}{2pt}
\resizebox{\columnwidth}{!}{%
\begin{tabular}{@{}l c cccccccc@{}}
\toprule
& \textbf{LIBERO} & \multicolumn{8}{c}{\textbf{LIBERO-Plus}} \\
\cmidrule(lr){2-2}\cmidrule(l){3-10}
\textbf{Model} & \textbf{Avg.} & \textbf{Cam.} & \textbf{Rob.} & \textbf{Lang.} & \textbf{Light} & \textbf{Bg.} & \textbf{Noise} & \textbf{Lay.} & \textbf{Avg.} \\
\midrule
\rowcolor{baselinebg}{RGB-only} & \textbf{98.9} & 64.1 & 86.5 & 96.2 & 98.4 & 80.7 & 82.5 & 83.4 & 84.0 \\
\midrule
$\lambda_{\mathrm{count}}=0.01$ & 98.6 & 61.8 & 87.4 & 96.0 & 98.4 & 80.1 & 82.9 & 84.0 & 83.9 \\
$\lambda_{\mathrm{count}}=0.03$ & 98.7 & 63.9 & 87.2 & \textbf{98.1} & \textbf{98.5} & \textbf{80.9} & 81.8 & 84.1 & \textbf{84.4} \\
$\lambda_{\mathrm{count}}=0.1$  & 98.7 & \textbf{65.5} & \textbf{88.8} & 95.1 & 94.5 & 78.1 & 81.9 & \textbf{84.3} & 83.8 \\
$\lambda_{\mathrm{count}}=0.3$  & 98.6 & 65.2 & 86.8 & 94.9 & 95.8 & 80.0 & \textbf{85.8} & 83.8 & 84.3 \\
$\lambda_{\mathrm{count}}=1$    & 98.3 & 65.2 & 86.0 & 94.1 & 96.1 & 77.0 & 84.5 & 82.6 & 83.3 \\
\bottomrule
\end{tabular}}
\end{table}

\paragraph{Event loss weight.}
With four bins, $\lambda_{\mathrm{count}}=0.03$ achieves the highest corrected LIBERO-Plus success in Table~\ref{tab:supp_res_lambda} (84.4\% vs.\ 84.0\% for RGB-only), followed by $\lambda_{\mathrm{count}}=0.3$ (84.3\%) and $\lambda_{\mathrm{count}}=0.01$ (83.9\%), with all weights within 0.7~pp of RGB-only.
The effects vary across perturbations, with sensor-noise success improving by 3.3~pp at $\lambda_{\mathrm{count}}=0.3$, while lighting and language success fall below those of RGB-only at the same weight.

\begin{table}[t]
\centering
\caption{\textbf{Temporal resolution of the residual event target.}
Success rates (\%) on LIBERO and corrected LIBERO-Plus with $\lambda_{\mathrm{count}}=0.03$; bins per 0.8\,s latent group, bin lengths in parentheses. Cam.: camera; Rob.: robot initial states; Lang.: language; Light: lighting; Bg.: background; Noise: sensor noise; Lay.: object layouts; Avg.: trial-weighted mean.
Bold: best per column.}

\label{tab:supp_res_bin}
\footnotesize
\setlength{\tabcolsep}{2pt}
\resizebox{\columnwidth}{!}{%
\begin{tabular}{@{}l c cccccccc@{}}
\toprule
& \textbf{LIBERO} & \multicolumn{8}{c}{\textbf{LIBERO-Plus}} \\
\cmidrule(lr){2-2}\cmidrule(l){3-10}
\textbf{Event target} & \textbf{Avg.} & \textbf{Cam.} & \textbf{Rob.} & \textbf{Lang.} & \textbf{Light} & \textbf{Bg.} & \textbf{Noise} & \textbf{Lay.} & \textbf{Avg.} \\
\midrule
\rowcolor{baselinebg}{RGB-only} & 98.9 & 64.1 & 86.5 & 96.2 & 98.4 & 80.7 & 82.5 & 83.4 & 84.0 \\
\midrule
1 bin (0.8\,s)  & \textbf{99.0} & 62.8 & 87.4 & 97.0 & 97.7 & 76.7 & \textbf{82.9} & 83.9 & 83.7 \\
2 bins (0.4\,s) & \textbf{99.0} & 65.2 & 87.6 & 96.9 & 95.3 & 79.9 & 82.2 & 83.9 & 84.1 \\
4 bins (0.2\,s) & 98.7 & 63.9 & 87.2 & \textbf{98.1} & \textbf{98.5} & \textbf{80.9} & 81.8 & \textbf{84.1} & \textbf{84.4} \\
8 bins (0.1\,s) & \textbf{99.0} & \textbf{66.0} & \textbf{87.8} & 95.1 & 97.2 & 77.8 & 81.8 & 84.0 & 83.9 \\
\bottomrule
\end{tabular}}
\end{table}

\paragraph{Temporal resolution.}
At $\lambda_{\mathrm{count}}=0.03$, corrected LIBERO-Plus success (Table~\ref{tab:supp_res_bin}) gradually increases from 83.7\% with one bin to 84.1\% with two bins and 84.4\% with four bins (0.2\,s each, matching our joint representation), before decreasing to 83.9\% with eight bins.
These results suggest modest benefits from finer temporal granularity up to four bins, without further improvement at higher resolution.

\paragraph{Prediction quality trade-off.}
We further evaluate RGB and event prediction quality across event loss weights on 32 recorded successful policy trajectories.
As shown in Table~\ref{tab:supp_res_quality}, increasing $\lambda_{\mathrm{count}}$ from 0.01 to 1.0 improves event activity F1 from 77.94\% to 86.56\% and reduces event log-count MAE from 0.7332 to 0.4742, while RGB PSNR decreases from 21.844 to 20.911\,dB.
At $\lambda_{\mathrm{count}}=0.03$, RGB PSNR remains comparable to RGB-only (21.808 vs.\ 21.805\,dB), with an event activity F1 of 81.14\%.
These results indicate a trade-off between RGB and event prediction quality as $\lambda_{\mathrm{count}}$ increases.

\begin{table}[t]
\centering
\caption{\textbf{Prediction quality of residual-aware supervision.}
RGB and event prediction metrics on 32 recorded successful policy trajectories using single-prefill prediction. 
Higher PSNR, SSIM, and event activity F1 and lower event log-count MAE indicate better quality.}
\label{tab:supp_res_quality}
\footnotesize
\setlength{\tabcolsep}{7pt}
\resizebox{\columnwidth}{!}{%
\begin{tabular}{lcccc}
\toprule
& \multicolumn{2}{c}{\textbf{RGB}} &
\multicolumn{2}{c}{\textbf{Event}} \\
\cmidrule(lr){2-3}\cmidrule(l){4-5}
\textbf{Method}
& \textbf{PSNR $\uparrow$}
& \textbf{SSIM $\uparrow$}
& \textbf{F1 $\uparrow$}
& \textbf{Log MAE $\downarrow$} \\
\midrule
\rowcolor{baselinebg}{RGB-only}
& 21.805 & \textbf{0.8302} & -- & -- \\
$\lambda_{\mathrm{count}}=0.01$
& \textbf{21.844} & 0.8290 & 77.94 & 0.7332 \\
$\lambda_{\mathrm{count}}=0.03$
& 21.808 & 0.8266 & 81.14 & 0.6403 \\
$\lambda_{\mathrm{count}}=1.0$
& 20.911 & 0.7896 & \textbf{86.56} & \textbf{0.4742} \\
\bottomrule
\end{tabular}}
\end{table}

Overall, residual-aware supervision demonstrates the feasibility of incorporating event-derived targets into RGB future modeling without modifying the inference pathway.
However, its policy gains remain modest and configuration-dependent.
On corrected LIBERO-Plus, the best residual variant achieves 84.4\%, compared with 84.0\% for RGB-only and 85.9\% for joint RGB--event modeling (Tables~\ref{tab:supp_res_lambda} and~\ref{tab:supp_flow_comparison}). Additionally, the prediction-quality analysis reveals a trade-off between RGB and event prediction as the auxiliary loss weight increases, although its implications for policy performance remain unclear.
These results favor the joint formulation in our setting, while highlighting the feasibility and limitations of residual-aware supervision.


\section{Additional Analysis}

\subsection{Future Prediction Quality under Perturbations}
\label{sec:supp_pred_perturb}

We extend the future-prediction evaluation in Sec.~4.6 of the main paper to five LIBERO-Plus perturbation categories.
We evaluate the same RGB-only, Event-only, and Joint models on 99 windows from 32 recorded policy trajectories, using three shared noise realizations per window and the same metric aggregation.
Future targets are rendered under the corresponding perturbations, except for sensor noise, which uses clean futures.
We exclude robot initial-state and object-layout perturbations, as the altered trajectories no longer match the recorded futures.

\begin{table}[t]
\centering
\caption{\textbf{Future prediction quality under LIBERO-Plus perturbations.}
H1 and H8 correspond to 0.2\,s and 1.6\,s futures.
Results use one video flow-velocity evaluation (NFE=1).
``Only'' refers to RGB-only for RGB prediction and Event-only for event prediction.}
\label{tab:supp_pred_quality}
\footnotesize
\setlength{\tabcolsep}{5pt}
\resizebox{\columnwidth}{!}{%
\begin{tabular}{@{}lllrrrr@{}}
\toprule
& & & \multicolumn{2}{c}{H1 (0.2\,s)} & \multicolumn{2}{c}{H8 (1.6\,s)} \\
\cmidrule(lr){4-5}\cmidrule(l){6-7}
Target & Metric & Condition & Only & Joint & Only & Joint \\
\midrule
\multirow{12}{*}{RGB}
 & \multirow{6}{*}{\shortstack[l]{PSNR $\uparrow$\\(dB)}}
   & Clean        & 27.02 & 26.88 & 21.26 & 21.26 \\
 & & Sensor noise & 23.09 & 23.00 & 19.80 & 19.65 \\
 & & Camera       & 23.62 & 23.69 & 18.66 & 18.69 \\
 & & Background   & 24.52 & 24.49 & 18.78 & 19.14 \\
 & & Lighting     & 27.44 & 27.76 & 19.57 & 19.96 \\
 & & Language     & 26.69 & 26.69 & 21.08 & 21.14 \\
\cmidrule(l){2-7}
 & \multirow{6}{*}{SSIM $\uparrow$}
   & Clean        & 0.867 & 0.864 & 0.780 & 0.781 \\
 & & Sensor noise & 0.788 & 0.786 & 0.731 & 0.727 \\
 & & Camera       & 0.816 & 0.819 & 0.703 & 0.704 \\
 & & Background   & 0.782 & 0.779 & 0.680 & 0.687 \\
 & & Lighting     & 0.882 & 0.880 & 0.768 & 0.777 \\
 & & Language     & 0.862 & 0.861 & 0.778 & 0.778 \\
\midrule
\multirow{6}{*}{Event}
 & \multirow{6}{*}{\shortstack[l]{Activity\\F1 $\uparrow$ (\%)}}
   & Clean        & 78.1 & 77.9 & 64.0 & 63.9 \\
 & & Sensor noise & 75.0 & 74.7 & 61.5 & 61.8 \\
 & & Camera       & 73.9 & 74.2 & 52.0 & 53.0 \\
 & & Background   & 67.3 & 68.2 & 52.1 & 54.4 \\
 & & Lighting     & 77.1 & 77.2 & 61.6 & 62.0 \\
 & & Language     & 77.7 & 77.7 & 62.6 & 63.5 \\
\bottomrule
\end{tabular}}
\end{table}

\begin{figure*}[tp]
\centering
{\small\bfseries Sensor noise\par}
\includegraphics[page=1,width=\textwidth]{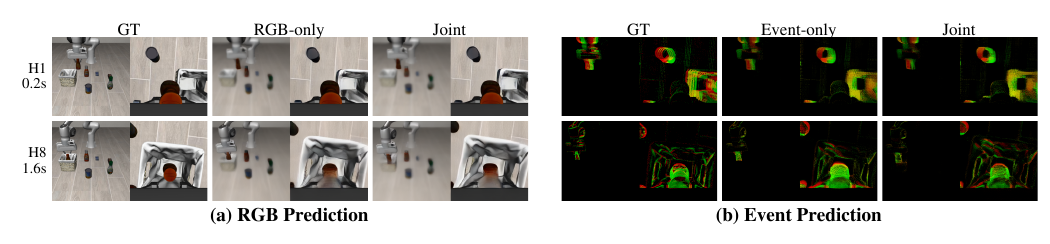}\par
\vspace{-3mm}
{\small\bfseries Camera\par}
\includegraphics[page=2,width=\textwidth]{figures/all7_perturbations_nfe1_main_format.pdf}\par
\vspace{-3mm}
{\small\bfseries Background\par}
\includegraphics[page=3,width=\textwidth]{figures/all7_perturbations_nfe1_main_format.pdf}\par
\vspace{-3mm}
{\small\bfseries Lighting\par}
\includegraphics[page=4,width=\textwidth]{figures/all7_perturbations_nfe1_main_format.pdf}\par
\vspace{-3mm}
{\small\bfseries Language\par}
\includegraphics[page=5,width=\textwidth]{figures/all7_perturbations_nfe1_main_format.pdf}
\caption{\textbf{Future predictions under LIBERO-Plus perturbations.}
(a) RGB: RGB-only vs.\ Joint.
(b) Event: Event-only vs.\ Joint.
H1/H8: 0.2/1.6\,s; NFE=1.
Head and wrist views are juxtaposed.
ON/OFF events are green/red.}

\label{fig:supp_pred_perturb}
\end{figure*}

Table~\ref{tab:supp_pred_quality} reports prediction quality at NFE=1, with qualitative comparisons in Fig.~\ref{fig:supp_pred_perturb}.
Across all conditions, prediction quality decreases from H1 to H8.
Compared with clean evaluation, visual perturbations reduce H8 RGB PSNR from approximately 21.3\,dB to 18.7--20.0\,dB, whereas language perturbations have little effect.
Sensor noise causes a larger PSNR reduction at H1 (approximately 3.9\,dB) than at H8 (1.5--1.6\,dB).
Overall, Joint and single-target models exhibit comparable prediction quality across perturbations.
At H8, Joint achieves modest improvements in RGB PSNR and SSIM under background and lighting changes, along with higher event activity F1 under all five perturbations.
Under sensor noise, Joint yields slightly lower PSNR and SSIM than RGB-only at both horizons.


Notably, prediction quality and policy success do not consistently follow the same trends.
Under sensor noise, Joint achieves higher success despite slightly lower reconstruction quality, whereas the opposite pattern is observed under lighting.
These observations suggest that aggregate prediction fidelity alone may be insufficient to characterize policy robustness.
However, since these metrics are evaluated on different samples, they do not establish a direct relationship between prediction quality and policy success.

\begin{figure*}[!htbp]
  \centering
  \includegraphics[width=\textwidth]{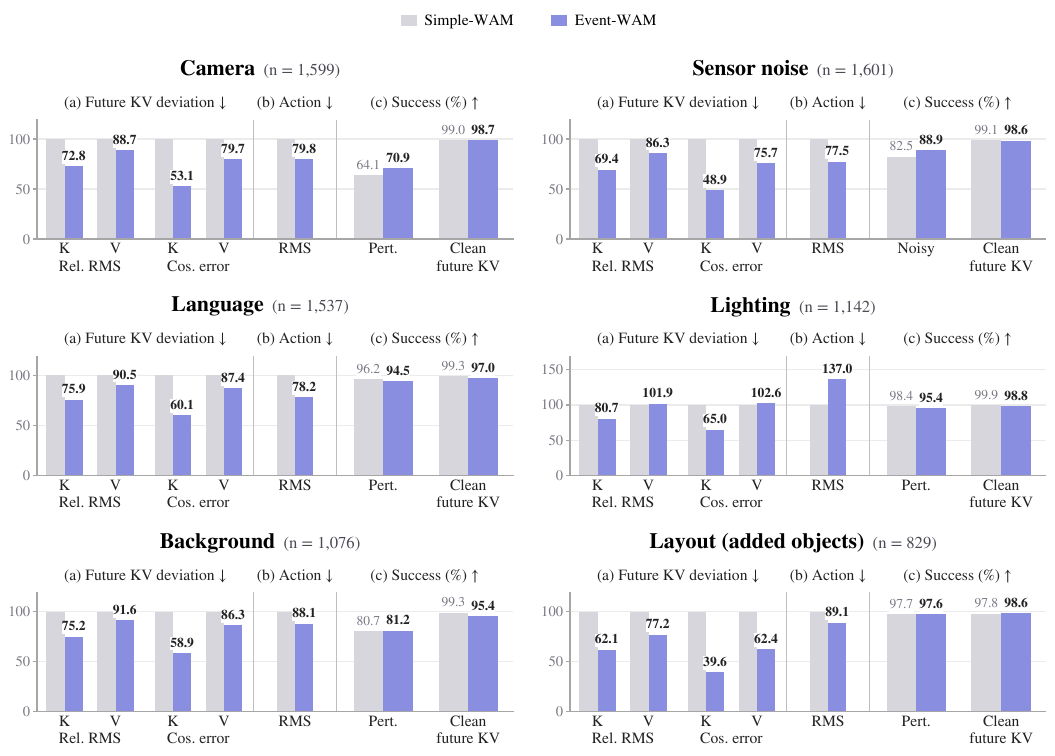}
  \caption{\textbf{Robustness of future KV under LIBERO-Plus perturbations.}
(a) Relative RMS, $\lVert\tilde X-X\rVert_F/\lVert X\rVert_F$, and cosine error, $1-\cos(\tilde X,X)$, of future keys (K) and values (V), where $X$ and $\tilde X$ denote each model's clean and perturbed features.
(b) RMS of perturbation-induced action changes through future KV over the first 10 executed actions.
(a,b) are normalized to Simple-WAM $=100$ within each perturbation group.
(c) Success rates (\%) without (Noisy or Pert.) and with clean future KV.
Lighting uses an extended vertical scale.
$n$: number of variants.
$\downarrow$/$\uparrow$: lower/higher is better.}
  \label{fig:supp-future-kv}
\end{figure*}

\subsection{Robustness of Future KV under Perturbations}
\label{sec:supp-future-kv}

We further extend the future-KV analysis in Fig.~7 of the main paper to additional LIBERO-Plus perturbations.
For each model, we compare future KV from perturbed and clean forward passes while keeping other inputs fixed.
The clean pass uses the unperturbed view or, for language, the canonical instruction.
For object-layout variants involving added distractors, we remove their visual appearance while preserving the underlying physical state.
We exclude robot initial-state and physical layout changes, as these alter the actions required by the task.
During action prediction, we replace only the future KV with its clean counterpart.
For language perturbations, we separately evaluate the effect of providing the canonical instruction only to the action model.

Under camera and sensor noise, Event-WAM achieves higher success than Simple-WAM (70.9\% vs.\ 64.1\% and 88.9\% vs.\ 82.5\%, respectively; Fig.~\ref{fig:supp-future-kv}(c)).
In both conditions, Event-WAM also exhibits smaller future-KV deviations and perturbation-induced action changes (Fig.~\ref{fig:supp-future-kv}(a,b)).
Replacing future KV with clean features raises both models' success rates to approximately 99\%, suggesting that future-KV corruption contributes substantially to failures under these perturbations.
Under lighting, however, Event-WAM exhibits larger future-V deviations and a 37.0\% greater action change, accompanied by lower success.
Taken together, these results suggest an association between future-KV stability and Event-WAM's gains under camera and sensor noise and its degradation under lighting.


\section{Real-World Experimental Details}
\label{sec:supp_real}

\subsection{Experimental Setup}
\label{sec:supp_real_setup}

\begin{table*}[t!]
\centering
\caption{\textbf{Real-world tasks.} Training and evaluation use the same instructions. Avg.\ len.\ is the average demonstration length.}
\label{tab:real_tasks}
\small
\setlength{\tabcolsep}{4pt}
\begin{tabularx}{\textwidth}{@{}l >{\raggedright\arraybackslash}X >{\raggedright\arraybackslash}X c c c@{}}
\toprule
Task & Instruction & Success criterion & Demos & Avg.\ len. & Limit \\
\midrule
Pen to cup   & ``Pick up the pen and place it in the cup.'' & The pen rests stably in the cup after release. & 100 & 13.1\,s & 30\,s \\
Gather dolls & ``Pick up all three dolls and place them in the basket.'' & All three dolls are in the basket. & 100 & 31.6\,s & 50\,s \\
Open drawer  & ``Open the drawer.'' & The drawer is fully open. & 100 & 8.8\,s & 30\,s \\
Stack cubes  & ``Pick up the red cube and stack it on the blue cube.'' & The red cube rests stably on the blue cube after release. & 100 & 11.6\,s & 30\,s \\
\bottomrule
\end{tabularx}
\end{table*}

\begin{figure}[t]
\centering
\includegraphics[width=\columnwidth]{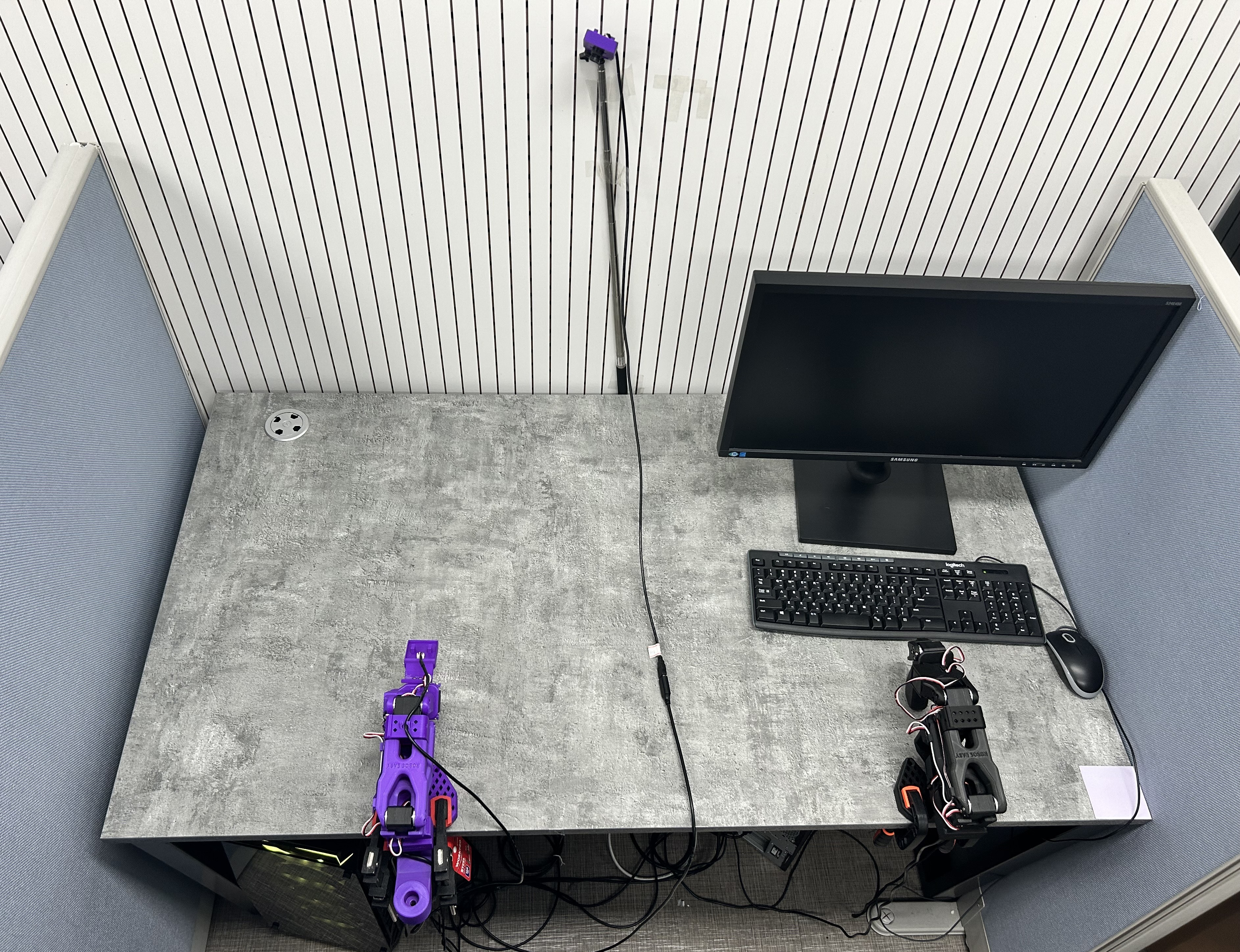}
\caption{\textbf{Real-world setup.} The SO-101 follower arm with the front and wrist RGB cameras on the tabletop workspace.}
\label{fig:real_setup}
\end{figure}

We use an SO-101 arm~\cite{so101} with five revolute joints and a gripper, observed by a front RGB camera and a wrist-mounted RGB camera.
The front camera is fixed across the table from the robot and views the workspace from above at an oblique angle.
Both cameras record $480\times640$ RGB images at 25\,Hz.
Each view is resized to $224\times224$, and the two views are concatenated horizontally into a $224\times448$ frame, as in simulation (Sec.~\ref{sec:supp_impl}).
States and actions are six-dimensional (five arm joints and the gripper), and actions are absolute joint-position targets.
All experiments are conducted on a tabletop in an indoor laboratory (Fig.~\ref{fig:real_setup}).

Table~\ref{tab:real_tasks} summarizes the language instruction, success criterion, and time limit for each task.
\emph{Pen to cup} requires grasping a narrow pen near its center of mass so that it stays balanced when placed into a cup.
\emph{Gather dolls} requires placing three differently colored dolls in a basket in any order.
\emph{Open drawer} requires inserting the gripper into a loop handle before pulling the drawer fully open.
\emph{Stack cubes} requires grasping a red cube nearly as wide as the maximum gripper opening and placing it stably on a blue cube.

We collect 400 successful demonstrations (100 per task) from four operators using leader--follower teleoperation with LeRobot.
Before each demonstration, task objects are placed at randomized positions and orientations within the reachable workspace.
Event targets are synthesized offline from both camera streams using the settings in Table~\ref{tab:supp_v2e}, with 0.16\,s bins and $\gamma_E=3.2189$.

We train a single language-conditioned policy across all four tasks for 30k steps, using a global batch size of 128 on four RTX~PRO~6000 GPUs.
All other hyperparameters follow Table~\ref{tab:supp_hparams}, including $\lambda_{\mathrm{RGB}}=\lambda_{\mathrm{Event}}=3$ and $p_1=0.5$.
The Simple-WAM baseline uses the same demonstrations, training schedule, and policy inputs.
Each training window contains 33 frames spanning 1.28\,s, with a corresponding 32-step action chunk and eight 0.16\,s event bins.
At inference, both methods run on a single RTX~3090 GPU, predicting 32-step action chunks with 10 action-denoising steps.
Actions are executed at a nominal control frequency of 25\,Hz, and a new chunk is requested every eight control steps (0.32\,s), when 24 of the 32 predicted actions remain.
As in simulation, the policy uses only RGB observations, proprioception, and the language instruction, without event inputs or decoded future representations.

\subsection{Evaluation}
\label{sec:supp_real_perturb}

\begin{table}[t]
\centering
\caption{\textbf{Real-world results.} Success rate (\%) with successful trials in parentheses. Trials are per method. Each perturbation uses 10 trials per task. $\Delta$: Event-WAM minus Simple-WAM, in points.}
\label{tab:real_results}
\footnotesize
\setlength{\tabcolsep}{0pt}
\begin{tabular*}{\columnwidth}{@{\extracolsep{\fill}}l c c c c@{}}
\toprule
Condition & Trials & Simple-WAM & Event-WAM & $\Delta$ \\
\midrule
\multicolumn{5}{@{}l}{\textit{Standard}} \\
\quad Pen to cup   & 50  & 72.0 (36)  & \textbf{82.0} (41)  & $+10.0$ \\
\quad Gather dolls & 50  & 60.0 (30)  & \textbf{72.0} (36)  & $+12.0$ \\
\quad Open drawer  & 50  & 48.0 (24)  & \textbf{74.0} (37)  & $+26.0$ \\
\quad Stack cubes  & 50  & 82.0 (41)  & \textbf{88.0} (44)  & $+6.0$ \\
\quad Overall      & 200 & 65.5 (131) & \textbf{79.0} (158) & $+13.5$ \\
\midrule
\multicolumn{5}{@{}l}{\textit{Perturbed}} \\
\quad Camera        & 40  & 35.0 (14) & \textbf{42.5} (17) & $+7.5$ \\
\quad Background    & 40  & 37.5 (15) & \textbf{55.0} (22) & $+17.5$ \\
\quad Sensor noise  & 40  & 22.5 (9)  & \textbf{42.5} (17) & $+20.0$ \\
\quad Object layout & 40  & 12.5 (5)  & \textbf{40.0} (16) & $+27.5$ \\
\quad Overall       & 160 & 26.9 (43) & \textbf{45.0} (72) & $+18.1$ \\
\bottomrule
\end{tabular*}
\end{table}
Each method is evaluated for 50 trials per task under standard conditions (200 trials) and 10 trials per task under each of four perturbations (40 trials per perturbation, 160 in total).
Initial object placements are predefined, marked on the table with transparent tape, and shared by both methods, which are evaluated alternately.
A trial ends when the success criterion in Table~\ref{tab:real_tasks} is met or the time limit expires.
Evaluation is not blinded; to limit subjective judgment, two observers jointly assess each trial using only the binary criteria in Table~\ref{tab:real_tasks}.
Neither method is fine-tuned or given additional data for perturbations.

The four perturbations are applied one at a time, and none appears in the demonstrations.
\emph{Camera}: the front camera is shifted a few centimeters to the left, tilted downward, and rotated toward the left; the wrist camera is unchanged.
\emph{Background}: the gray tabletop is covered with a white tablecloth.
\emph{Sensor noise}: Gaussian noise is added in software to every front-camera frame, leaving the wrist camera and the physical scene unchanged.
\emph{Object layout}: task-irrelevant distractor objects are added while the task objects keep their predefined placements.
Figure 2 of the main paper shows initial and final frames for one representative task under each perturbation.

As shown in Table~\ref{tab:real_results}, Event-WAM outperforms Simple-WAM on all four tasks under standard conditions (79.0\% vs.\ 65.5\%), with the largest gain on Open drawer (+26.0~pp), which requires precise handle insertion.
Under perturbations, Event-WAM also leads overall (45.0\% vs.\ 26.9\%), with the largest gain under object layout (+27.5~pp) and the smallest under camera shift (+7.5~pp).
Both overall differences are statistically significant (two-sided Fisher's exact test, $p<0.01$); per-task (50 trials) and per-perturbation (40 trials) differences should be read as trends.

\subsection{Qualitative Results}
\label{sec:supp_qual}

\begin{figure*}[!t]
\centering
\includegraphics[width=\textwidth]{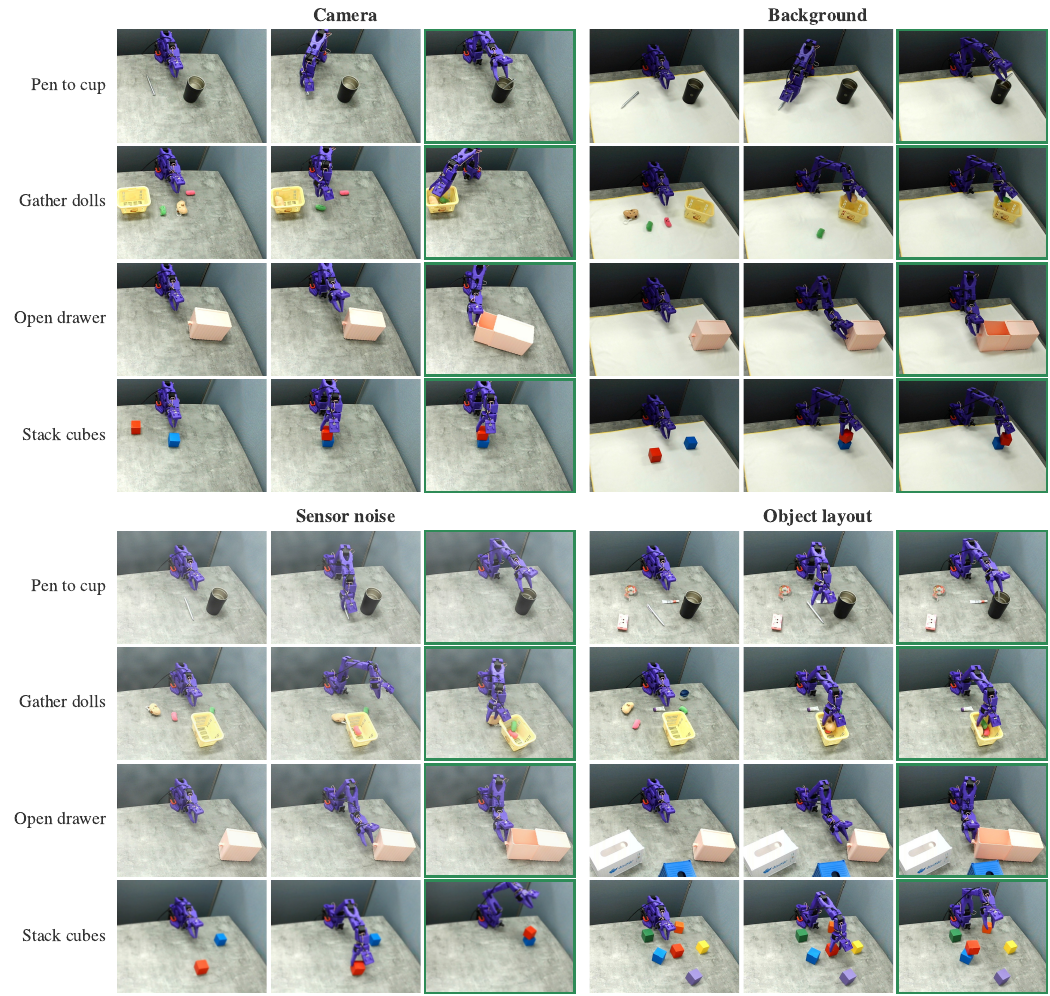}
\caption{\textbf{Successful Event-WAM rollouts across real-world perturbations.}
One successful trial is shown for each of the 16 perturbation--task combinations, with initial, intermediate, and final front-camera frames.}
\label{fig:real_grid}
\end{figure*}

\begin{figure*}[!t]
\centering
\includegraphics[width=\textwidth]{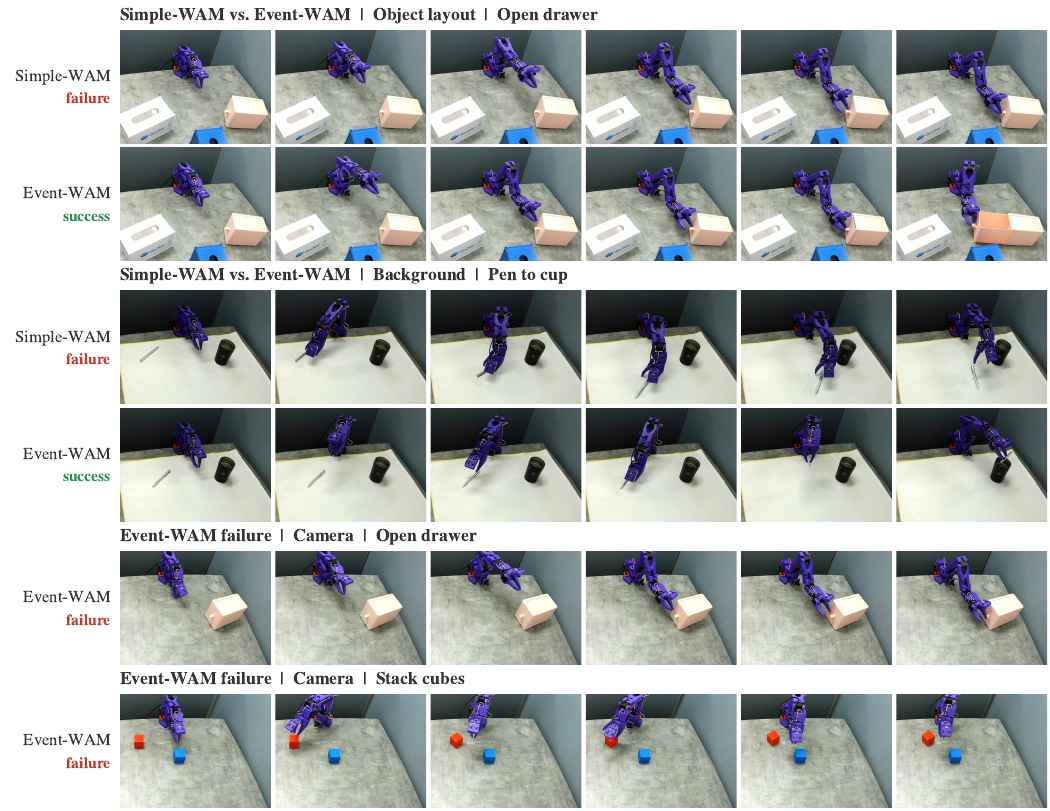}
\caption{\textbf{Real-world policy comparisons and failure cases.}
Top: Simple-WAM vs.\ Event-WAM on the same task and perturbation.
Bottom: Event-WAM failures under the camera perturbation.
Front-camera frames are sampled across each episode.}

\label{fig:real_compare}
\end{figure*}

Figure~\ref{fig:real_grid} illustrates successful Event-WAM rollouts across all 16 real-world perturbation--task combinations.
Figure~\ref{fig:real_compare} (top) compares the two policies on the same task under the same perturbation.
Under the object-layout perturbation, both policies reach the drawer, but only Event-WAM pulls it open.
Under the background perturbation, only Event-WAM grasps the pen and places it in the cup.

Figure~\ref{fig:real_compare} (bottom) shows two Event-WAM failures under the camera perturbation.
In Open drawer, the gripper reaches the drawer but pushes the box instead of hooking the handle.
In Stack cubes, the gripper repeatedly pushes the red cube instead of grasping it.
Both cases illustrate failures in precise manipulation despite approaching the target object, consistent with the relatively small improvement under the camera perturbation (Table~\ref{tab:real_results}).

\end{document}